\documentclass{article}

\usepackage[utf8]{inputenc}

\usepackage[preprint]{corl_2024} %

\usepackage{multicol}

\usepackage{amsmath}
\usepackage{amssymb}
\usepackage{booktabs}
\usepackage{tabularx}

\usepackage{bm}
\usepackage{booktabs}
\usepackage{multirow}

\usepackage{float}
\usepackage{caption} 
\usepackage{subcaption}
\usepackage{balance}

\usepackage{amsfonts}
\usepackage{algorithmic}
\usepackage{graphicx}
\usepackage{textcomp}
\usepackage{todonotes}
\usepackage{csquotes}
\usepackage{xspace}
\usepackage{siunitx}

\usepackage{longtable}
\usepackage{floatflt}

\usepackage{multicol}
\usepackage{multirow}

\usepackage{lipsum}

\usepackage[shortlabels, inline]{enumitem}
\usepackage{wrapfig}

\usepackage{pifont}
\usepackage{lipsum}  

\NewDocumentCommand{\rot}{O{45} O{1em} m}{\makebox[#2][l]{\rotatebox{#1}{#3}}}%

\newcommand{\ie}{\textrm{i.e.,}\xspace}

\newcommand{\etal}{\textrm{et~al.}\xspace}

\newcommand{\rlbench}{\text{RLBench}\xspace}

\newcommand{\peract}{\texttt{PerAct}\xspace}
\newcommand{\peracttwo}{\texttt{PerAct$^2$}\xspace}
\newcommand{\alohaact}{\texttt{ACT}\xspace}
\newcommand{\rvt}{\texttt{RVT}\xspace}
\newcommand{\rvtlf}{\texttt{RVT-LF}\xspace}
\newcommand{\peractlf}{\texttt{PerAct-LF}\xspace}
\newcommand{\leaderfollower}{\text{leader-follower}\xspace}

\newcommand{\xmark}{\includegraphics[width=0.45em]{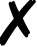}}
\newcommand{\arrowup}{\includegraphics[width=0.45em]{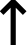}}
\newcommand{\arrowdown}{\includegraphics[width=0.45em]{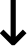}}
\newcommand{\starstar}{\includegraphics[width=0.75em]{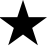}\xspace}

\usepackage{makecell}
\usepackage{ifthen} 
\newboolean{anonymized}
\setboolean{anonymized}{false}

\graphicspath{{images/}}

\title{\peracttwo: Benchmarking and Learning for \\  Robotic Bimanual Manipulation Tasks}

\author{Markus Grotz \\
University of Washington  \\
\texttt{grotz@cs.washington.edu}
\And
Mohit Shridhar \\
University of Washington, Dyson RLL \\
\texttt{mshr@cs.washington.edu}
\AND
Tamim Asfour \\
Karlsruhe Institute of Technology \\
\texttt{asfour@kit.edu}
\And
Dieter Fox \\
University of Washington, NVIDIA \\
\texttt{fox@cs.washington.edu}
}

\begin{document}
\maketitle

\begin{abstract}
Bimanual manipulation is challenging due to precise spatial and temporal coordination required between two arms.
While there exist several real-world bimanual systems, there is a lack of simulated benchmarks with a large task diversity for systematically studying bimanual capabilities across a wide range of tabletop tasks.
This paper addresses the gap by 
extending \rlbench~\cite{james2020rlbench} to bimanual manipulation. 
We open-source our code and benchmark, which comprises 13 new tasks with 23 unique task variations, each requiring a high degree of coordination and adaptability. 
To kickstart the benchmark, we extended several state-of-the-art methods to bimanual manipulation and also present a language-conditioned behavioral cloning agent -- \peracttwo, an extension of the \peract~\cite{shridhar2022} framework. This method enables the learning and execution of bimanual 6-DoF manipulation tasks. Our novel network architecture efficiently integrates language processing with action prediction, allowing robots to understand and perform complex bimanual tasks in response to user-specified goals.
The project website with code is available at: 
\url{http://bimanual.github.io}
\end{abstract}


\section{Introduction}

\begin{wrapfigure}{r}{0.5\textwidth}
    \vspace{-\normalbaselineskip}
    \centering
    \includegraphics[width=.48\textwidth,trim={0cm 0cm 0 0cm},clip]{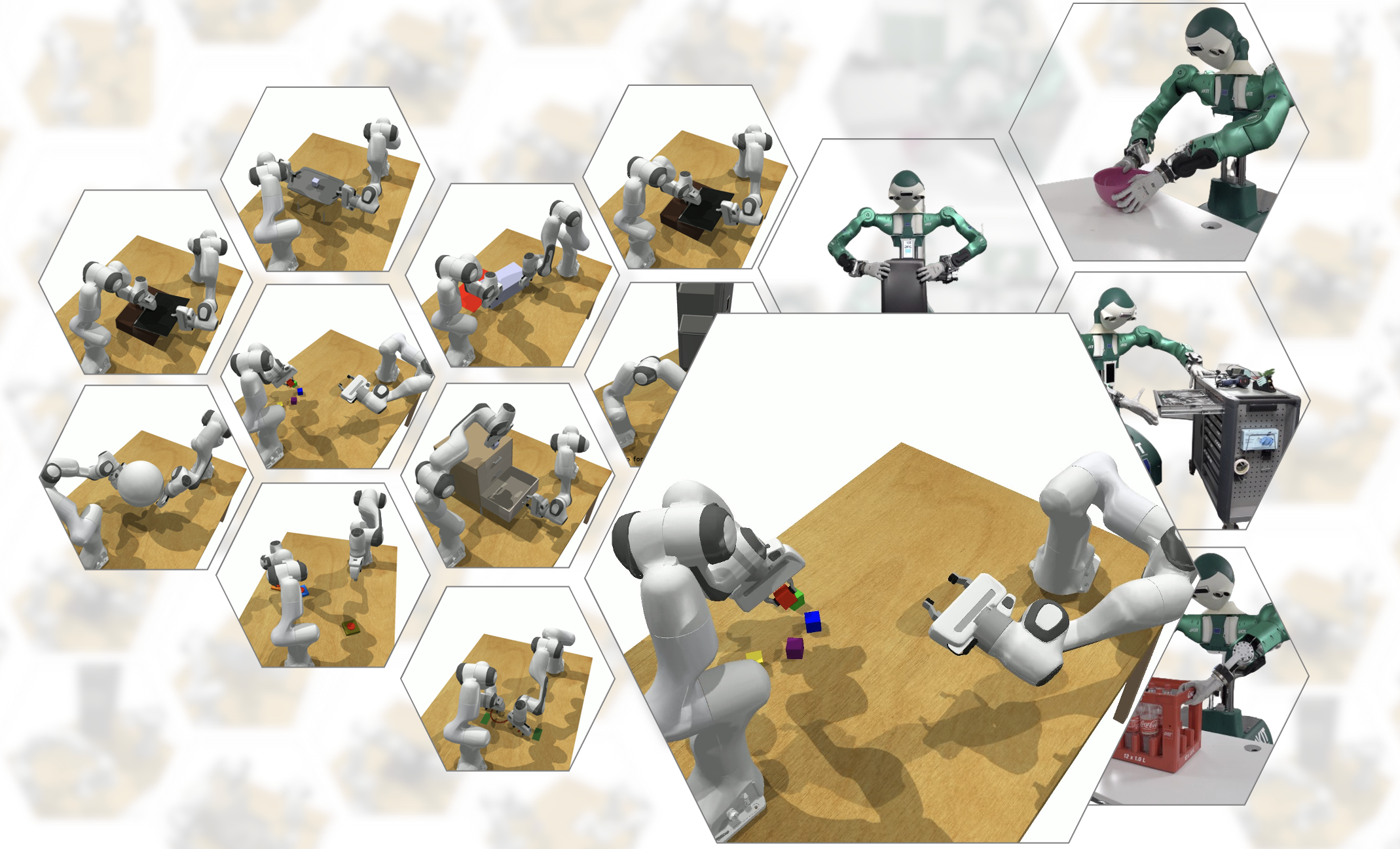}\hfill
    \caption{Selected bimanual tasks from the benchmark as well as real-world examples. 
    Due to the architecture design the method can easily be transferred to other robots as the policy outputs a 6-D pose and is agnostic to the underlying controller.}
    \label{fig:teaser}
\end{wrapfigure}

Humans seamlessly manipulate and interact with their environment using both hands. 
With both hands, humans achieve greater efficiency through enhanced reachability and can solve more sophisticated tasks.
Despite the recent advances in grasping and manipulation planning \cite{billard2019trends, newbury.2022} the investigation of bimanual 
manipulation remains an under-explored area, especially in terms of learning a manipulation policy. 
Unlike tasks that require grasping or manipulation with a single hand, bimanual manipulation introduces a layer of complexity due to the need for spatial and temporal coordination and a deep understanding of the task at hand. %
This complexity is compounded by the dynamic nature of real-world tasks, where the state of the environment and the objects within it are constantly changing, demanding continuous adjustment and coordination between both arms.

With the recent advent of complex bimanual systems such as the Boston Dynamics' Atlas, Tesla's Optimus or Figure AI's Humanoid, experiments investigating bimanual manipulation in real-world tasks are rising. 
Notably, the work by Zhao \etal \cite{zhao2023learning, fu2024mobile}
presents sophisticated and fine-grained real-world tasks learned from demos collected by teleoperation. 
While real-world tasks provide a rich context for understanding the challenges of bimanual manipulation, they suffer from issues of reproducibility and variability that make systematic assessment difficult. 
To advance research in bimanual manipulation, there is a critical need for a dedicated and rich bimanual benchmark that allows for the reproducible and systematic evaluation of new methods and models.
To fill this gap, we expand the robot learning benchmark \rlbench~\cite{james2020rlbench} to bimanual manipulation.
Moreover, we extend two existing unimanual learning-based methods, namely \peract and \rvt, to bimanual manipulation and compare those with \alohaact.
While benchmarking, we found that running two separate agents is insufficient and that coordination is a crucial aspect. Hence, we present a method to learn bimanual actions as well coordination implicitly using language-conditioned behavior-cloning agent within a single network. 
Overall, our contributions can be summarized as follows:
\begin{enumerate}[label=\arabic*.), topsep=0pt] %
    \item A benchmark with 13 bimanual manipulation tasks and 23 unique tasks variations. \rlbench is used as a basis preserving its functionality and its key properties.
    \item A novel network architecture, called \peracttwo, based on the \peract framework to predict bimanual manipulation actions, and
    \item Qualitative experiments in real world.
\end{enumerate}

We acknowledge the complexity of the tasks included in the benchmarks and look forward to the research community to embrace these challenges. We also hope that our method and evaluation will greatly enhance the benchmarking and generalization of skill learning in bimanual robots, including humanoids.

\section{Related Work}

\paragraph{Benchmarking}

\begin{wraptable}{r}{0.56\textwidth}
    \vspace{-\normalbaselineskip}
    \centering
    \begin{tabular}{@{}lccc@{}}
        \toprule
        \makecell{Benchmark \\ Name}        & \# \makecell{Bimanual \\ Tasks\footnotemark[1]} & \makecell{Task \\ Variation} & \makecell{Dataset \\ Generation} \\ 
        \midrule

        Robotsuite  & 3 & \checkmark & \xmark\footnotemark[2] \\
        ManiSkill   & 2 & \checkmark & \xmark \\
        RLBench     & -- & \checkmark & \checkmark \\
        Orbit       & 1 & \checkmark & \xmark\footnotemark[2] \\
        \makebox[0em]{\hspace{6.5em}  HumanoidBench} & 8 & \checkmark & \xmark \\
        ours        & 13 & \checkmark & \checkmark  \\ 
        \bottomrule
    \end{tabular}
    \caption{Overview of bimanual benchmarks.}
    \label{tab:related_work}
\end{wraptable}

\footnotetext[1]{We only count tasks were both arms are required.}
\footnotetext[2]{Robomimic allows for dataset generation either through human demonstrations or with a baseline.}

Benchmark protocols and frameworks for robotic manipulation are designed around reproducibility and extensibility.
For reinforcement learning, Fan \etal~\cite{fan18a} introduce \textit{SURREAL} to foster reproducibility for learning robotic manipulation tasks. 
Similarly, \textit{robosuite} \cite{robosuite2020}, \textit{bulletarm} \cite{wang2022bulletarm} and \textit{ManiSkill2}  \cite{gu2023maniskill2} provide a standardized benchmark and  learning-environment for robotic manipulation. While some of them have three (\textit{robosuite}) or two (\textit{ManiSkill2}) bimanual tasks, those are not sufficient to efficiently evaluate methods for bimanual manipulation.
James \etal~\cite{james2020rlbench} present \rlbench, a large scale benchmark and learning-environment for robot learning alongside with baseline algorithms. %
A crucial aspect here is the automated waypoint-based dataset generation, removing the need of
human demonstrations or by another baseline such as in \cite{mandlekar2021matters}.
However, no bimanual manipulation tasks were considered in \rlbench.
An issue that arises is the comparability, especially, in real world scenarios.  RB2 \cite{dasari2022rb2} aims to provide  rankings for robotic manipulation tasks by pooling data across labs.
Mittal \etal~\cite{mittal2023orbit} introduce \textit{Orbit}, a framework with GPU acceleration and photo realistic scenes, to  tackle real-to-sim gap. Their work also includes a bimanual manipulation task, but this was not the main focus of the work.
Other works have a focus on dexterous bimanual hand-object manipulation, such as \cite{fan2023arctic} or \cite{chen_towards_2022}.  However, these works focus on the hand and neglect the use of a robotic arm.
Last but not least, instead of providing a set of standardized benchmarks in simulation,
another way is to establish protocols for real-world robot experiments.
Chatzilygeroudis \etal~\cite{chatzilygeroudis2020benchmark} outline a protocol for bimanual manipulation of two challenging tasks for semi-deformable objects. 
Recently, \cite{sferrazza2024humanoidbench} introduced \textit{HumanoidBench}, a benchmark for whole-body manipulation and locomotion for reinforcement learning. 
\autoref{tab:related_work} overviews different robotic benchmarks.

\paragraph{Bimanual Manipulation}
Bimanual manipulation offers several advantages, such as increased reachability and enhanced dexterity \cite{smith2012dual}.  In general approaches vary depending on the domain. Key challenges for bimanual manipulation are the coordination and the state complexity, \ie how to orchestrate the arms with respect to each other. In the following, we want to specifically discuss work that addresses those issues.
\\
Coordination is central aspect for bimanual manipulation, for example when a robot is playing the piano \cite{robopianist2023}.
Coordination can be achieved by modeling coordination constraints explicitly besides task constraints \cite{sina_mirrazavi_salehian_coordinated_2016} to reach a moving target simultaneously.
With TAMP this robotic assembly planning can be  solved explicitly \cite{hartmann_long-horizon_2023}. Other areas include object carrying \cite{sirintuna_carrying_2023}.
Indeed, this requires knowledge about the environment and the physical parameters, which can be cumbersome when generalizing to other tasks and scenarios.
With an explicit \leaderfollower assumption and when a DMP \cite{dmp} for the leader is known, the coordination can be also learned using a structured-transformer to generate DMPs for the follower arm \cite{liu2022stirr}.
Other work \cite{zhao2023dualafford} learns a separate coordination module for each gripper and coordination is done by a separate module.
Concurrently, research by Grannen \etal investigates using one arm to stabilize an object while manipulating it with the other \cite{grannen2023stabilize}.
The symmetry-aware context has been studied for multi-object handover and rearrangement tasks \cite{li2023handover}. This work has also been evaluated in real-world.
Notably, for real-world robotics notably Zhao \etal \cite{zhao2023learning} learn bimanual manipulation from teleoperation. The work was later extended to mobile manipulation~\cite{fu2024mobile} and also revised~\cite{aloha2team2024aloha}.
In independent and concurrent work, \cite{liu2024} propose VoxActB, a voxel-based, language-conditioned method for bimanual manipulation using VLMs to focus on the most important regions. This work is complementary to ours, and future work could merge techniques from both works.

\vspace{-0.2cm}

\section{Method}
\label{sec:method}

\vspace{-0.2cm}

\begin{figure*}[tp] %
    \includegraphics[width=.98\textwidth,trim={0 9.1cm 2.45cm 0},clip]{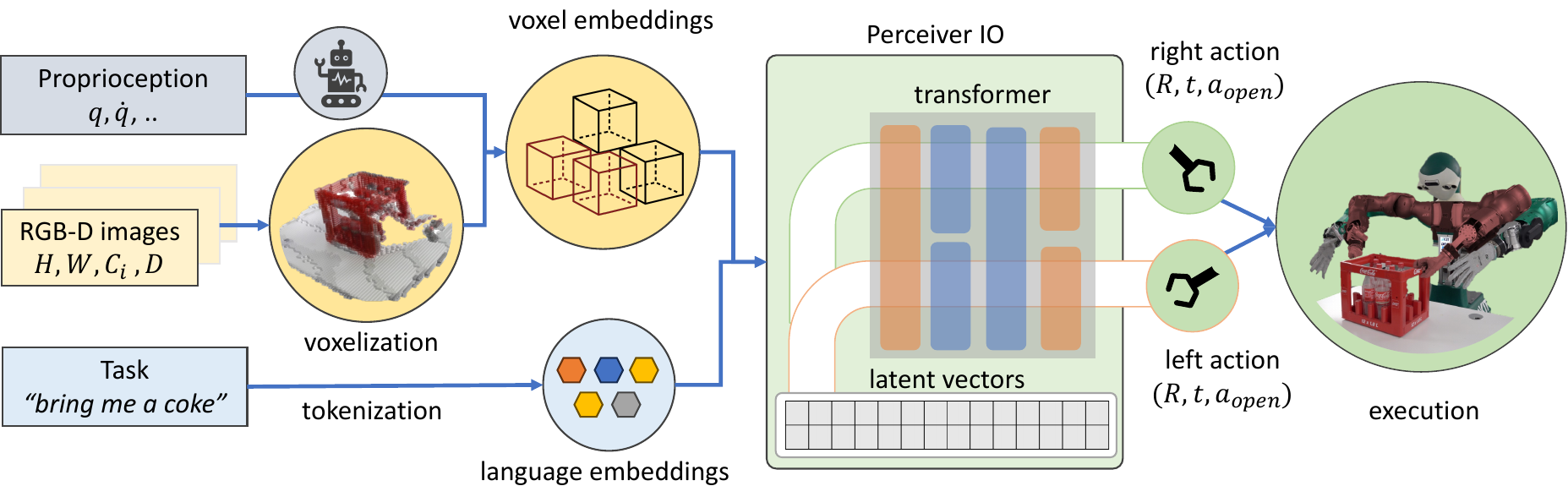}\hfill
    \caption{The system architecture. \peracttwo takes proprioception, RGB-D camera images as well as a task description as input. The voxel grid is constructed by merging data from multiple RGB-D cameras. A PerceiverIO transformer is utilized to learn features at both the voxel and language levels. The output for each robot arm includes a discretized action, which comprises a six-dimensional end-effector pose, the state of the gripper, and an extra indicator whether the motion-planner should use collision avoidance.}
    \label{fig:system_architecture}
\end{figure*}

To benchmark, we extend \rlbench \cite{james2020rlbench} to the complex bimanual case by adding functionality and tasks for bimanual manipulation. We choose \rlbench for several key advantages over other frameworks, mainly its ability to generate training data with variations as well as the widespread acceptance in the learning community.
To kickstart the benchmark, we also present a bimanual behavioral cloning agent.
Our method, called \peracttwo, extends \peract \cite{shridhar2022}, which learns a single language-conditioned policy for unimanual manipulation actions.  \autoref{fig:system_architecture} illustrates the system architecture, which has been modified to accommodate the intricate coordination required between two robotic arms while responding to language instructions.

\vspace{-0.1cm}

\subsection{RLBench2}

\vspace{-0.1cm}

\rlbench is a  robot learning benchmark suite featuring over more than 100 tasks to facilitate robot learning, which is widely used in the community. Among task diversity, other key properties include reproducibility and the ability to adapt to different learning strategies. 
We extend \rlbench to bimanual manipulation, while keeping the functionality and its key properties. This allows us to quantify the success of our method and compare it with other baselines. Compared to unimanual manipulation, bimanual manipulation is more challenging as it requires different kinds of coordination and orchestration of the two arms. 
Therefore, we also provide task descriptions along with metrics to benchmark performance and outline the key challenges of each task in \autoref{sec:description_of_tasks}.
For the implementation side this makes it much more complex since synchronization is required when controlling both arms at the same time.

\subsection{Task and Challenges}

\begin{figure}[bt]
    \centering
    \captionsetup[subfigure]{justification=centering}
    \begin{subfigure}[b]{.195\textwidth}
        \centering
        \includegraphics[width=\textwidth]{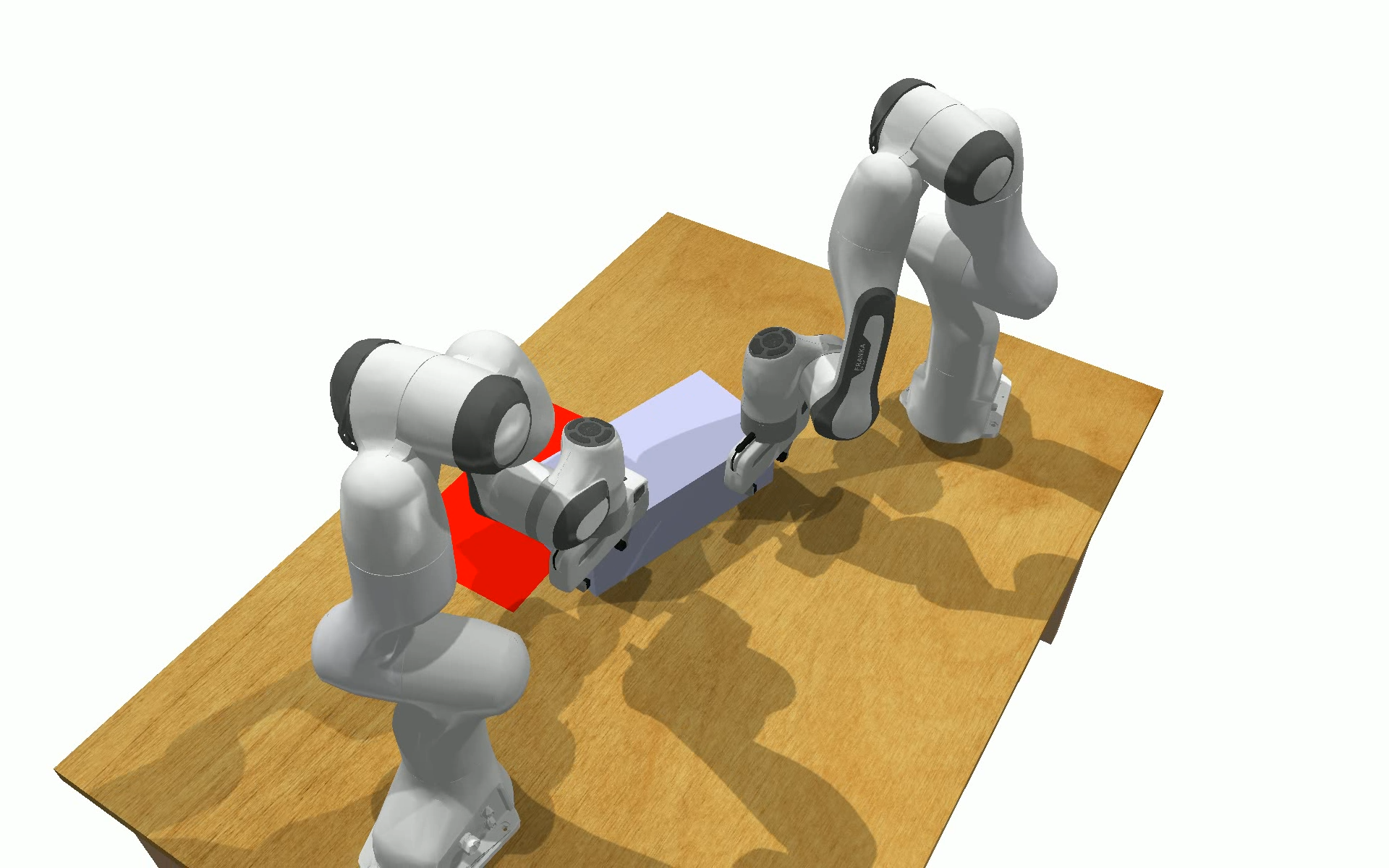}
        \caption{push box\\\,}
    \end{subfigure}%
    \begin{subfigure}[b]{.195\textwidth}
        \centering
        \includegraphics[width=\textwidth]{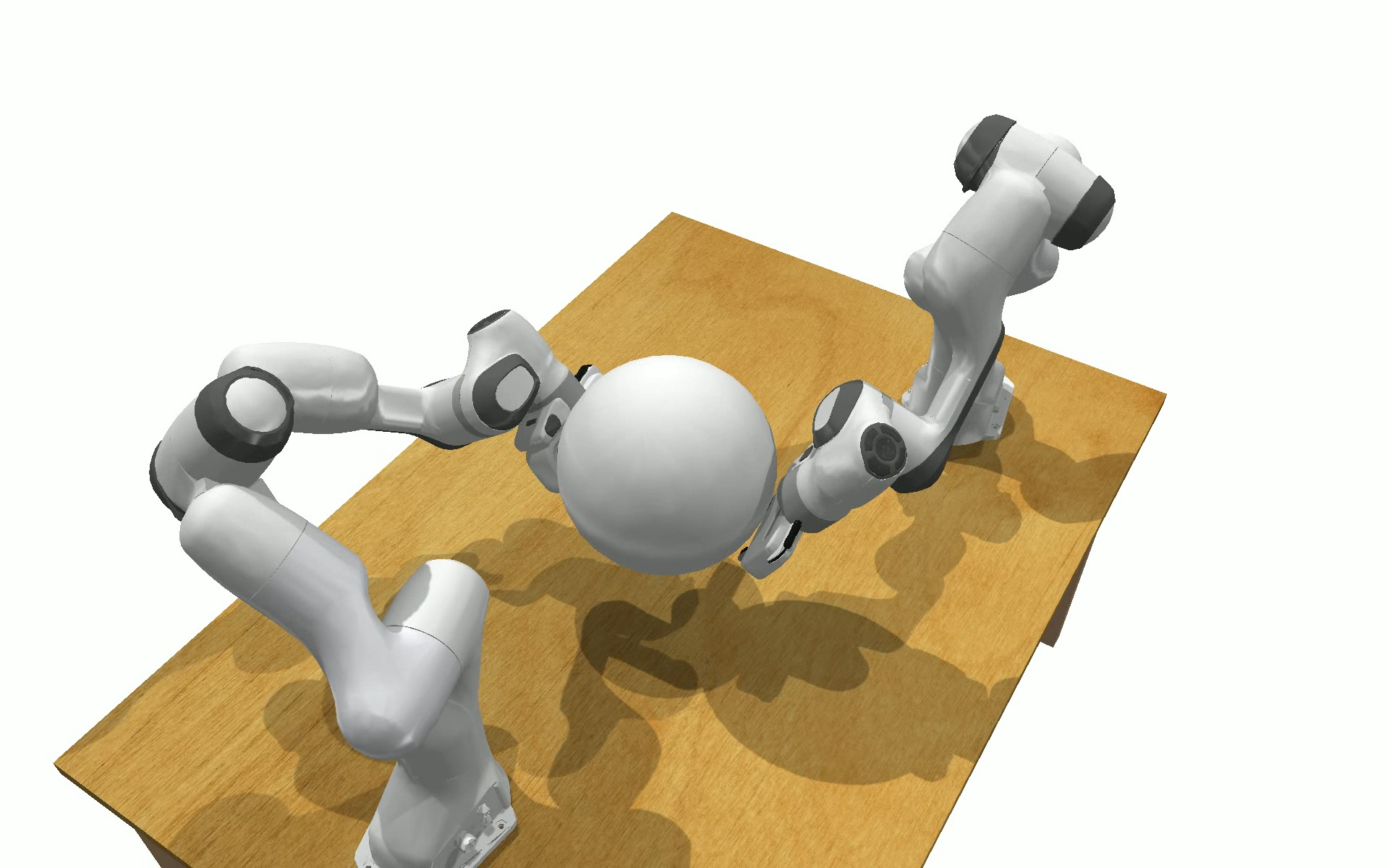}
        \caption{lift a ball\\\,}
    \end{subfigure}%
    \begin{subfigure}[b]{.195\textwidth}
        \centering
        \includegraphics[width=\textwidth]{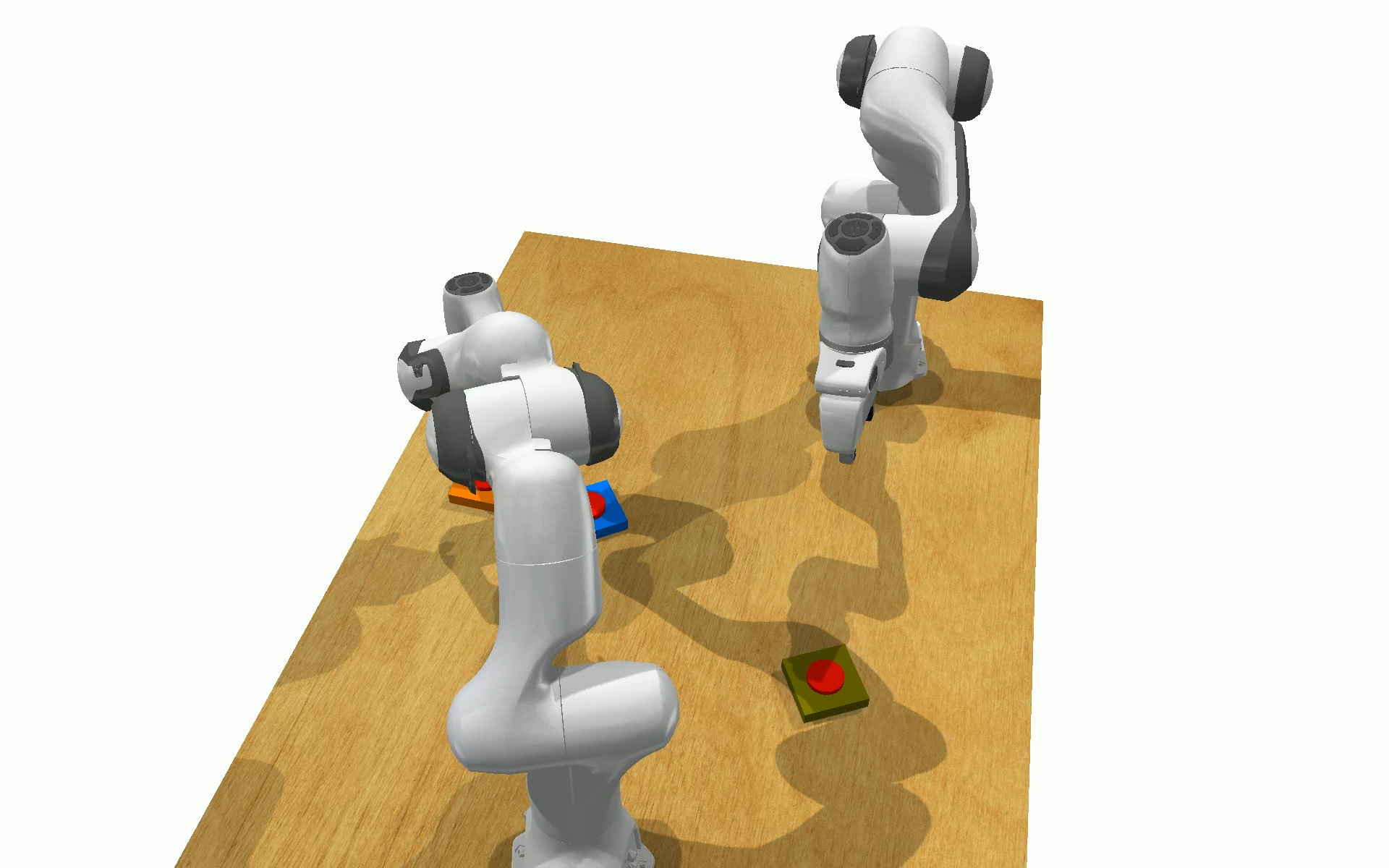}
        \caption{press two buttons\\\,}
    \end{subfigure}%
    \begin{subfigure}[b]{.195\textwidth}
        \centering
        \includegraphics[width=\textwidth]{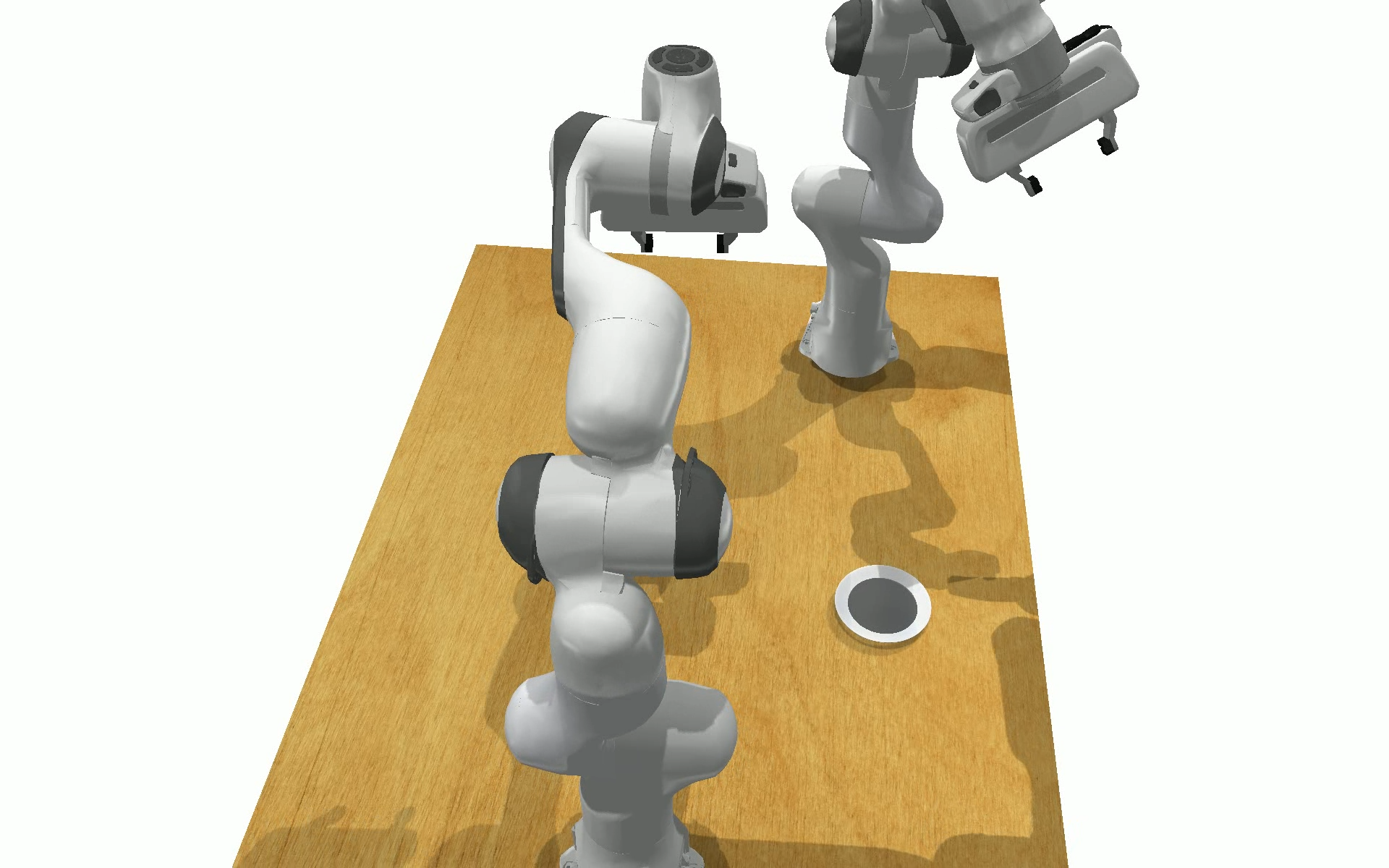}
        \caption{pick up a plate\\\,}
    \end{subfigure}%
    \begin{subfigure}[b]{.195\textwidth}
        \centering
        \includegraphics[width=\textwidth]{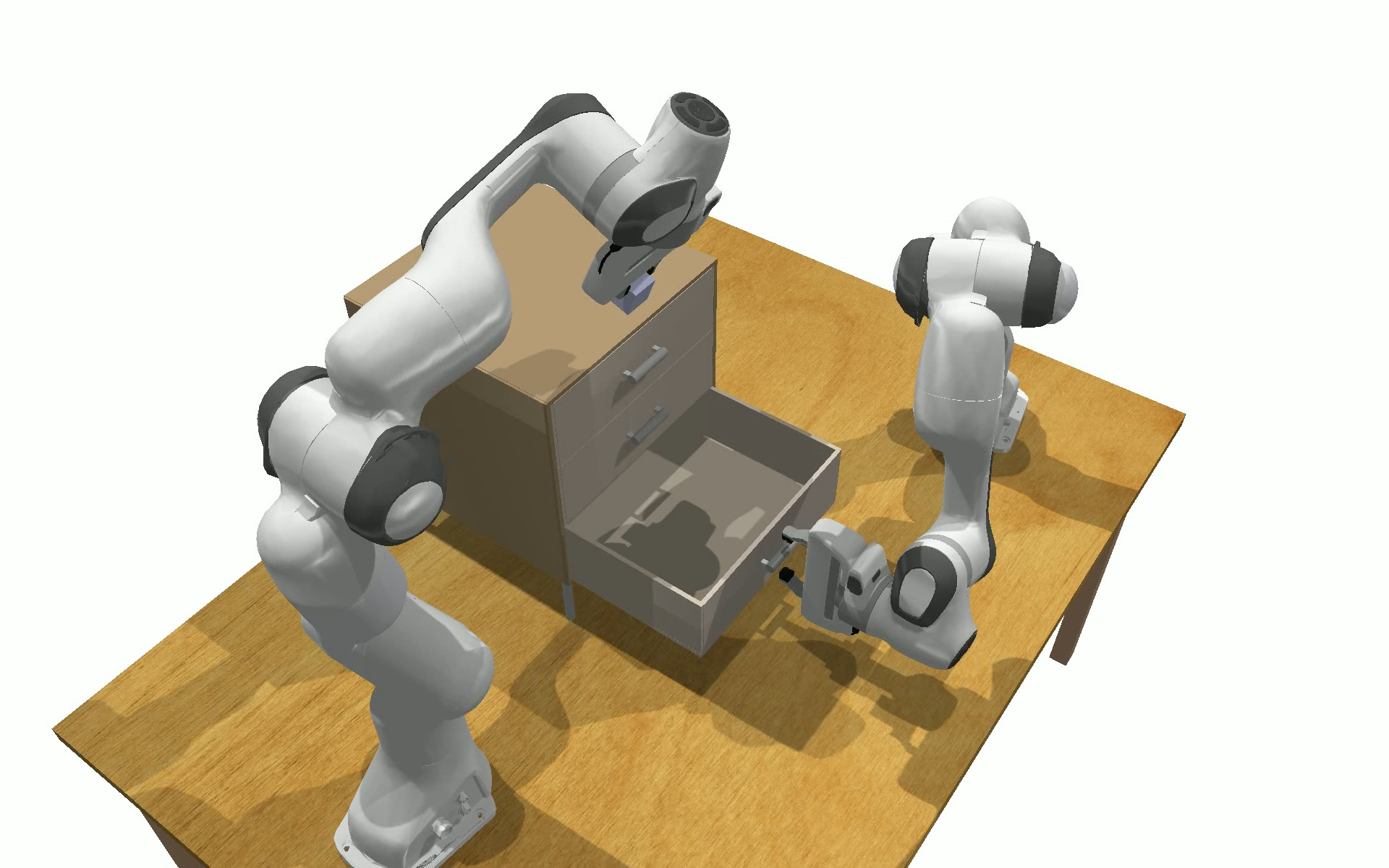}
        \caption{put item in drawer}
    \end{subfigure}
    
    \vspace{1.0\baselineskip}
    
    \begin{subfigure}[b]{.195\textwidth}
        \centering
        \includegraphics[width=\textwidth]{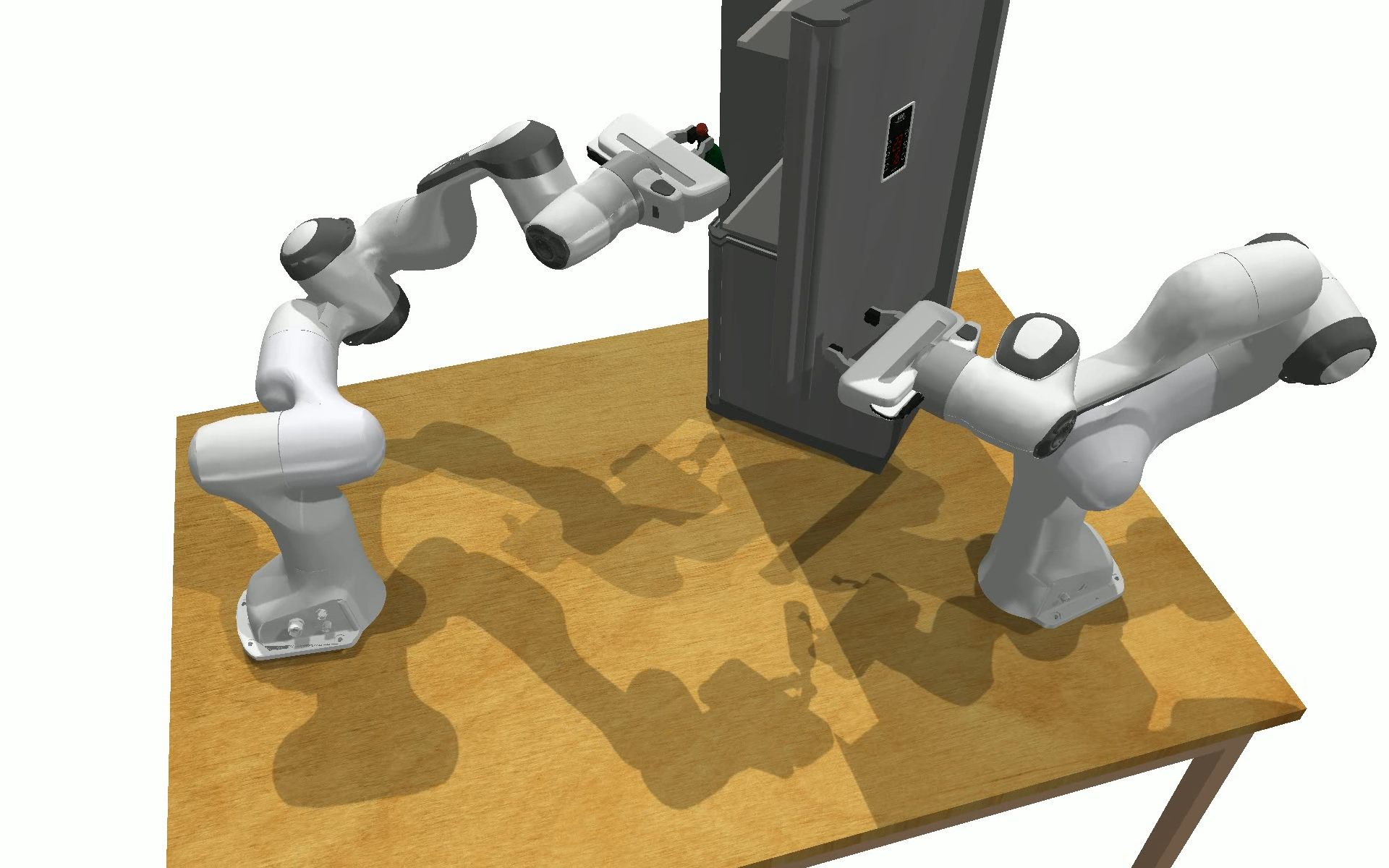}
        \caption{put bottle in fridge}
    \end{subfigure}%
    \begin{subfigure}[b]{.195\textwidth}
        \centering
        \includegraphics[width=\textwidth]{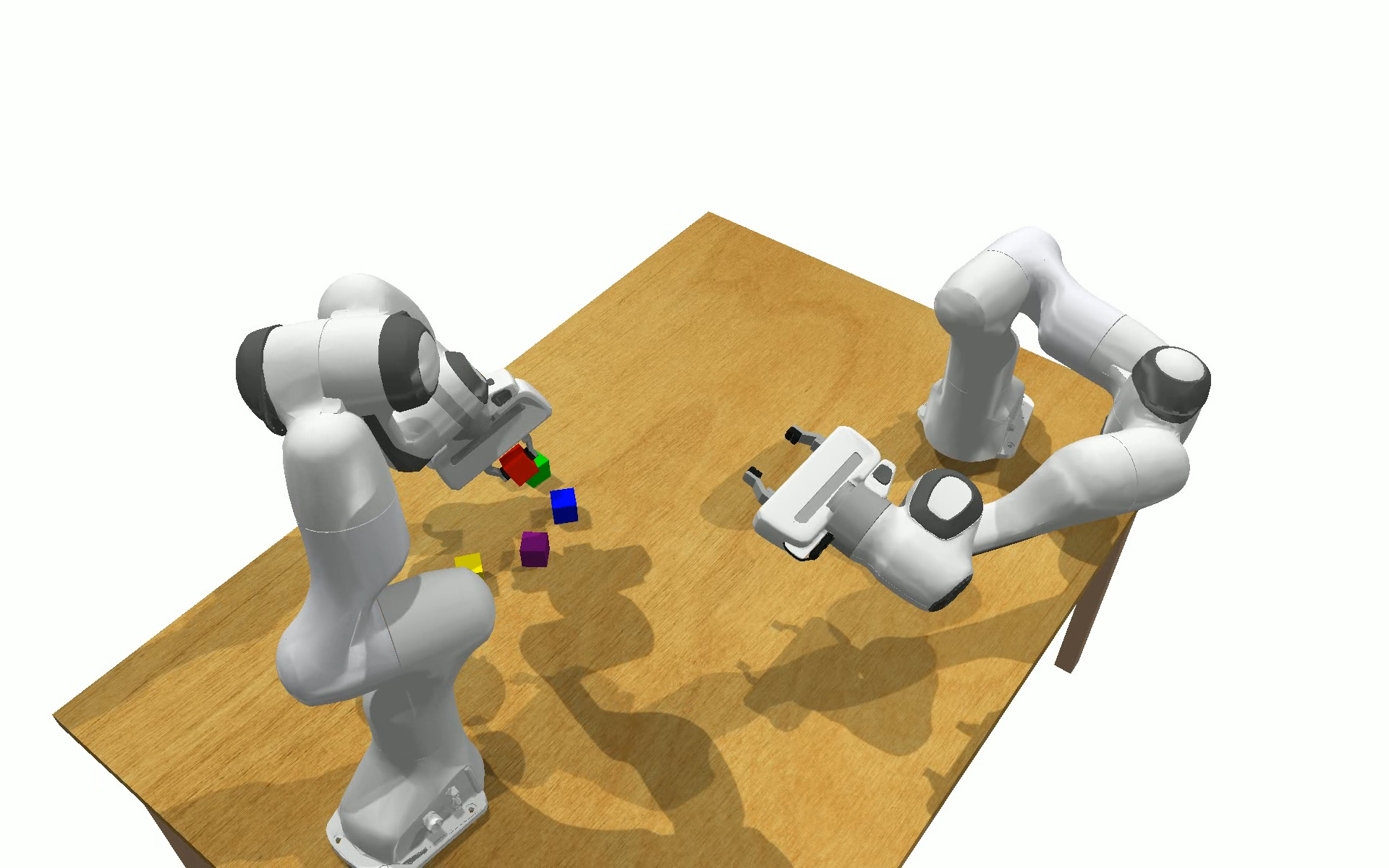}
        \caption{handover \\ an item}
    \end{subfigure}%
    \begin{subfigure}[b]{.195\textwidth}
        \centering
        \includegraphics[width=\textwidth]{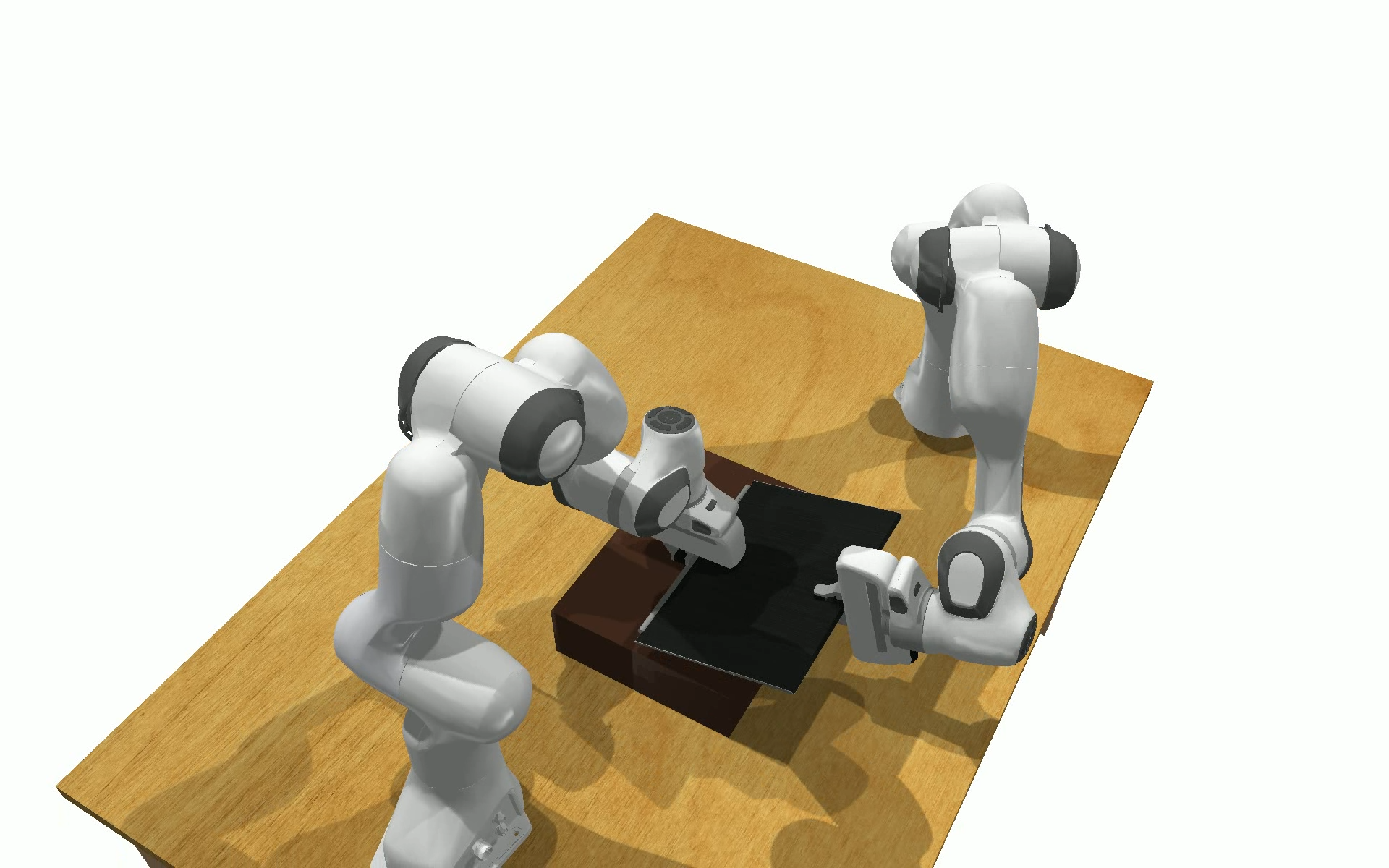}
        \caption{pick up notebook\\\,}
    \end{subfigure}%
    \begin{subfigure}[b]{.195\textwidth}
        \centering
        \includegraphics[width=\textwidth]{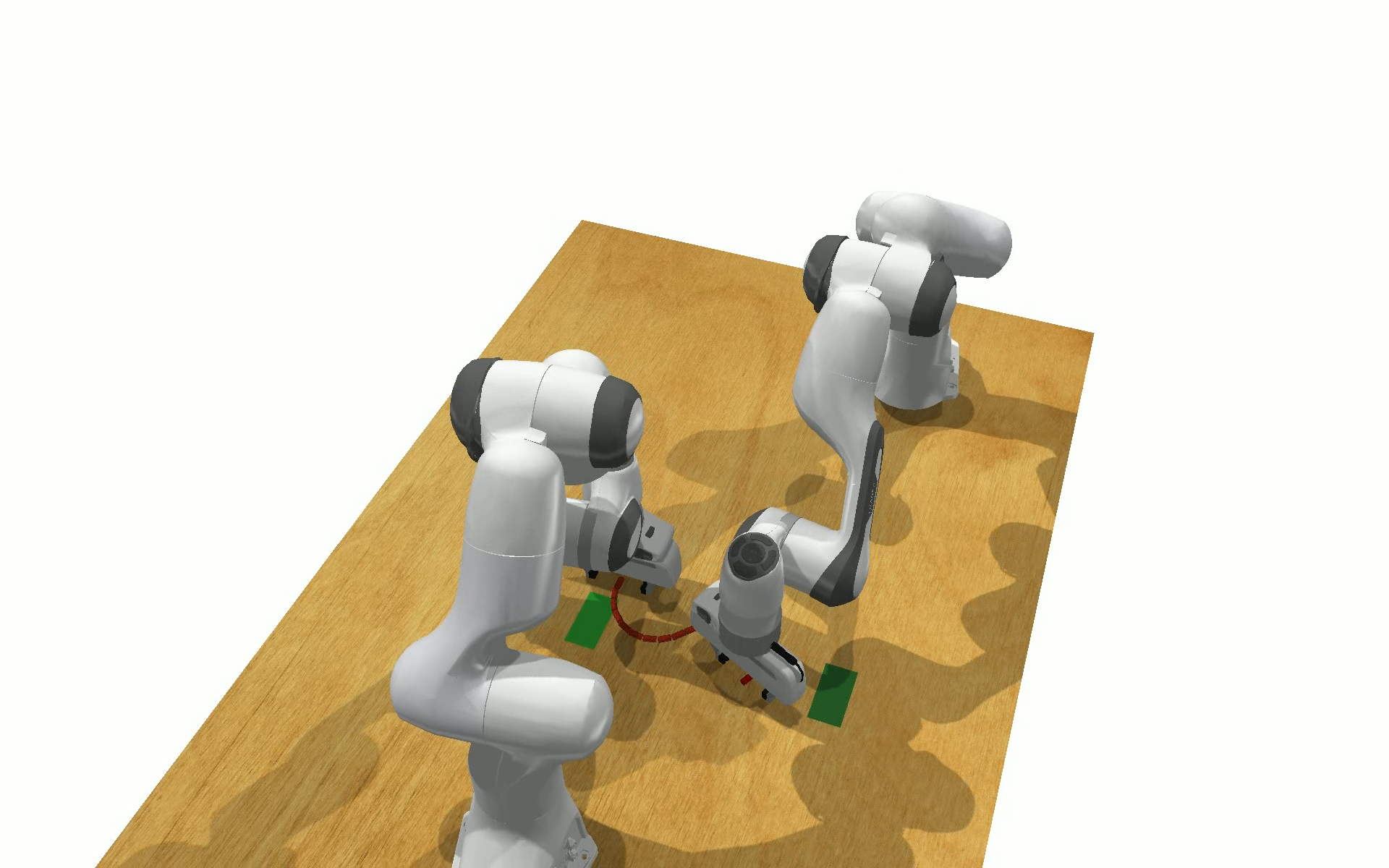}
        \caption{straighten rope\\\,}
    \end{subfigure}%
    \begin{subfigure}[b]{.195\textwidth}
        \centering
        \includegraphics[width=\textwidth]{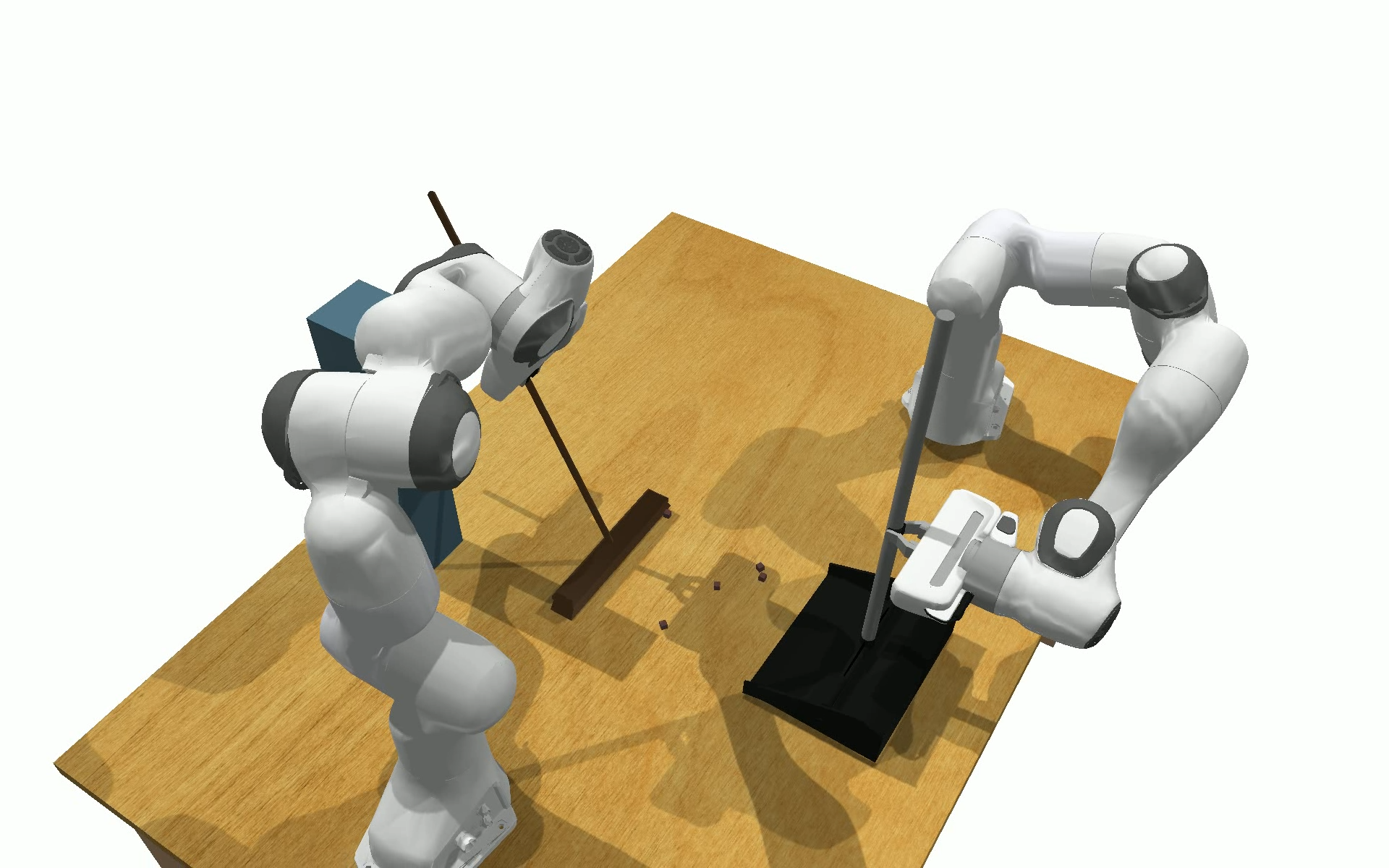}
        \caption{sweep dustpan\\\,}
    \end{subfigure}
    
    \vspace{0.5\baselineskip}
    
    \begin{subfigure}[b]{.195\textwidth}
        \centering
        \includegraphics[width=\textwidth]{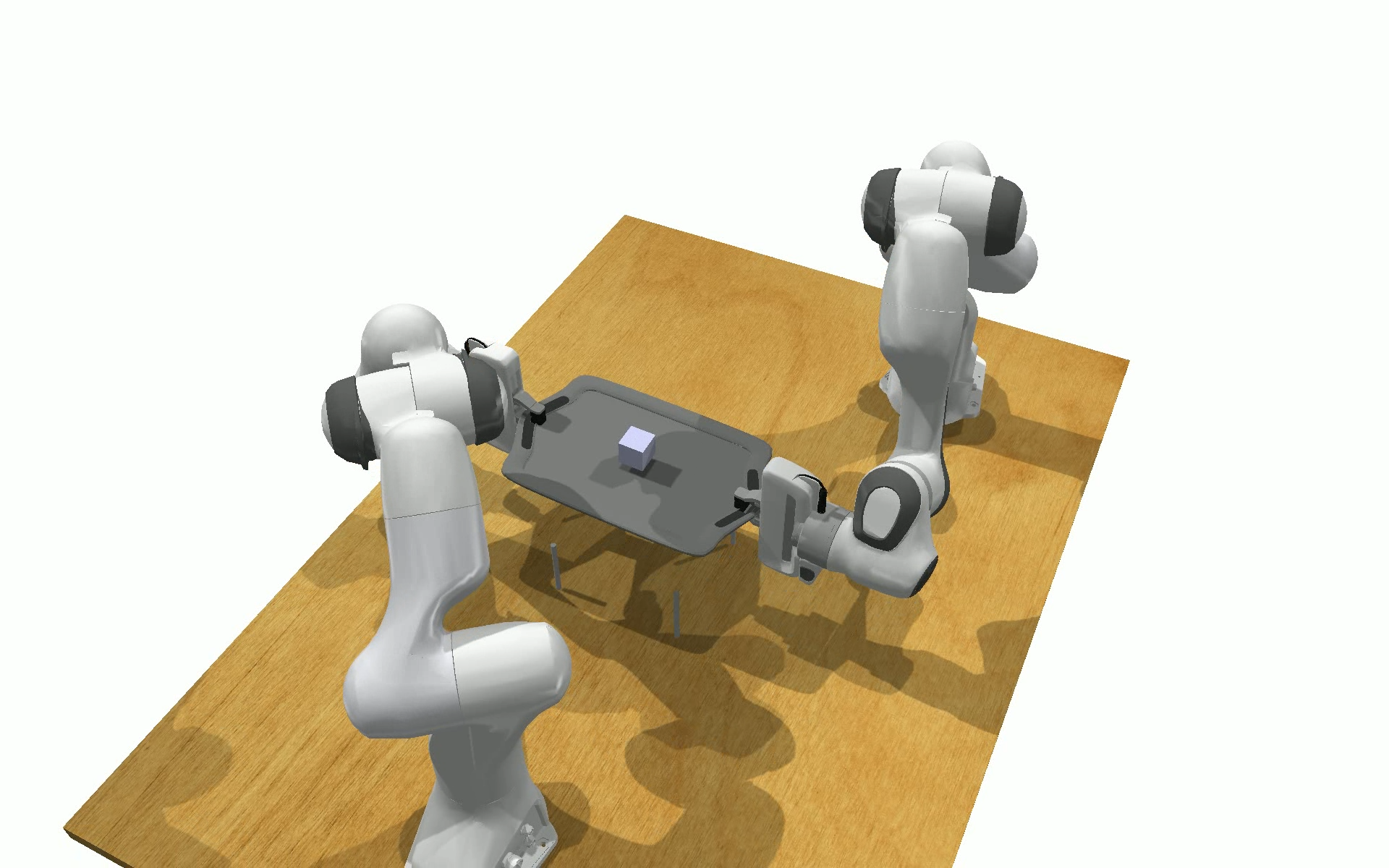}
        \caption{lift tray\\\,}
    \end{subfigure}%
    \begin{subfigure}[b]{.195\textwidth}
        \centering
        \includegraphics[width=\textwidth]{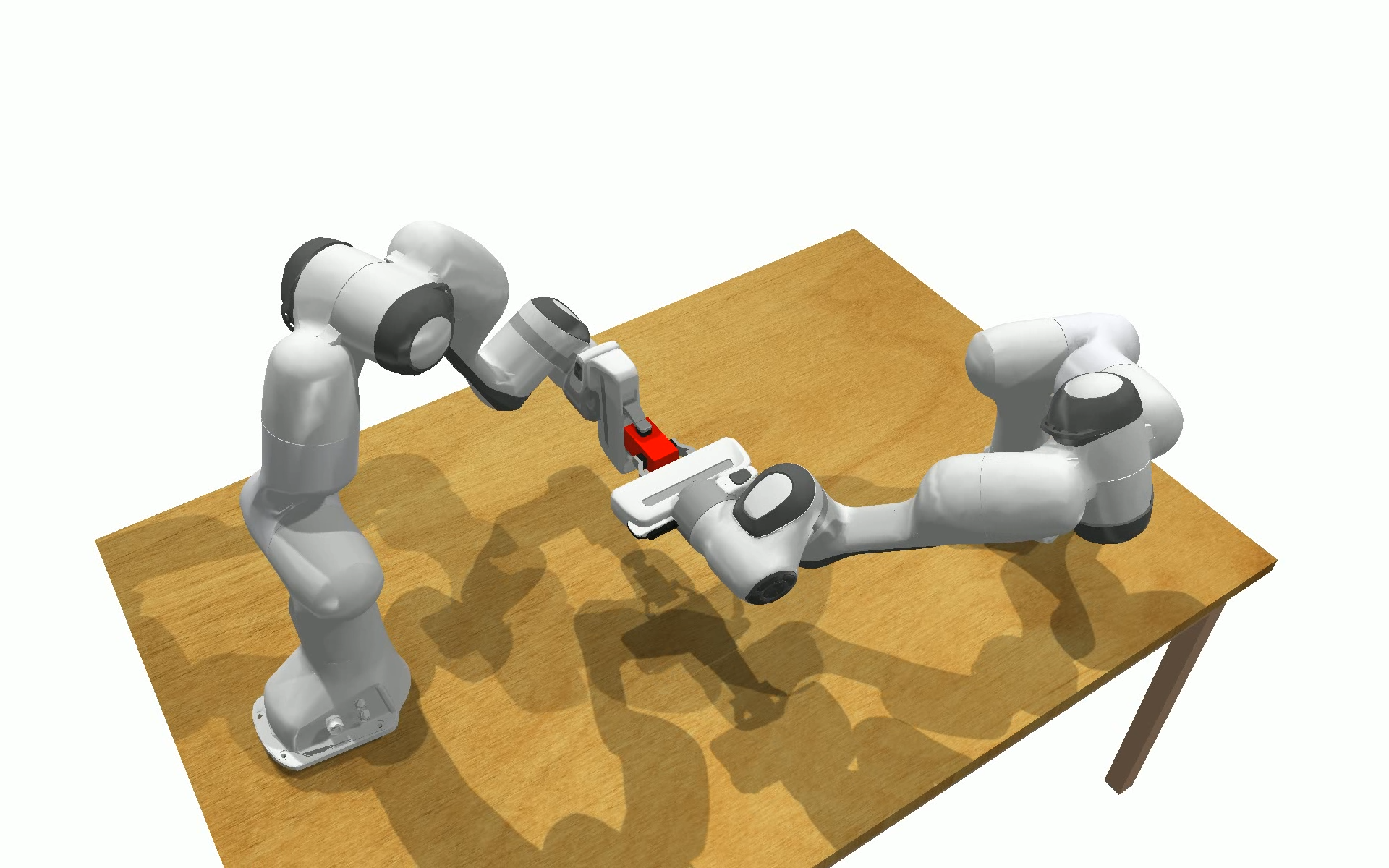}
        \caption{handover an item (easy)}
    \end{subfigure}%
    \begin{subfigure}[b]{.195\textwidth}
        \centering
        \includegraphics[width=\textwidth]{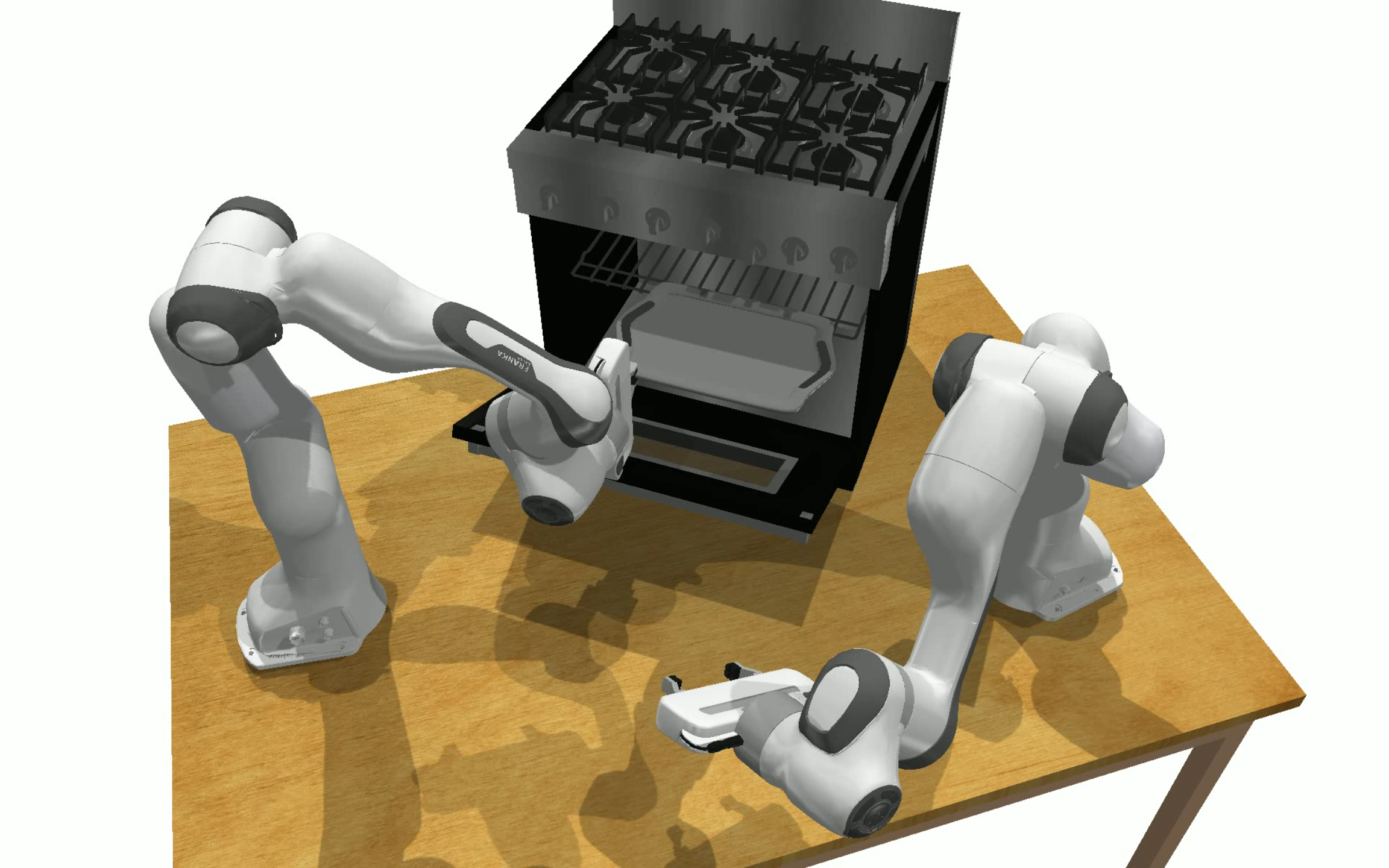}
        \caption{take tray out of oven}
    \end{subfigure}
    \vspace{0.5\baselineskip}
    \caption{Overview of the different tasks. For example, the task visualized (g) includes a handover of a specific item.}
    \label{fig:task_overview}
\end{figure}

\begin{table}[t]
\centering
\begin{tabular}{@{}lccccccc@{}}
\toprule
\multicolumn{1}{c}{\multirow{3}{*}{Task}} & \multicolumn{3}{c}{Coupled}          & \multicolumn{2}{c}{Coordination} \\ \cmidrule(l){2-4} \cmidrule(l){5-6}
\multicolumn{1}{c}{}                      & temporal   & spatial    & physical   & sym.      & sync.   \\ \midrule
(a) push box                              & \checkmark & \checkmark & \xmark     & \checkmark     & \checkmark      \\
(b) lift a ball                           & \checkmark & \checkmark & \checkmark & \checkmark     & \checkmark      \\
(c) push two buttons                      & \checkmark & \xmark     & \xmark     & \checkmark     & \xmark          \\
(d) pick up a plate                       & \checkmark & \checkmark & \checkmark & \xmark         & \xmark          \\
(e) put item in drawer                    & \checkmark & \xmark     & \xmark     & \xmark         & \xmark          \\
(f) put bottle in fridge                  & \checkmark & \xmark     & \xmark     & \xmark         & \xmark          \\
(g) handover an item                      & \checkmark & \checkmark & \checkmark & \xmark         & \checkmark      \\
(h) pick up notebook                      & \checkmark & \checkmark & \checkmark & \xmark         & \xmark          \\
(i) straighten rope                       & \checkmark & \checkmark & \checkmark & \xmark         & \checkmark      \\
(j) sweep dustpan                        & \checkmark & \checkmark & \checkmark & \xmark         & \xmark          \\
(k) lift tray                             & \checkmark & \checkmark & \checkmark & \checkmark     & \checkmark      \\
(l) handover item (easy)                  & \checkmark & \checkmark & \checkmark & \xmark         & \xmark          \\ 

(m) take tray out of oven                 & \checkmark & \xmark     & \xmark     & \xmark         & \xmark          \\
\bottomrule
\end{tabular}%
\vspace{0.5\normalbaselineskip}
\caption{Properties of the bimanual tasks. We report on the average length of the task demonstration in seconds. The average number of extracted keyframes of the task, the number of items that the robot can interact with and the task variations.}
\label{tab:task_overview}
\end{table}

We introduce 13 bimanual manipulation tasks with different coupling, coordination, language instructions and manipulation skills. \autoref{fig:task_overview} illustrates these tasks, which range from instructions like \textit{``push the box to the target area''} to  \textit{``put the bottle into the fridge''}. Out of the tasks, eight are prehensile manipulation, three are non-prehensile manipulation, and two involve both.  These tasks also show different kind of coupling and coordination. For example, the task \textit{``lift the tray''} has to be executed synchronously and both arms must be coordinated.
\autoref{tab:task_overview} classifies the tasks according to the bimanual taxonomy of \cite{Krebs2022}. Here, key distinguishing factors are the coupling as well as the required coordination between the two arms. We extended the classification in that we also distinguish between physical coupling, \ie if one arm exerts a force that could be measured by the other arm.
The benchmarking tasks differ in terms of complexity and the coordination required between two arms. Other attributes, such as the number of objects and the variation count, also affect the complexity of the task.
All of these attributes influence the task complexity, and a rich and diverse set of tasks is required for both qualitative and quantitative benchmarking.

\subsection{\peracttwo}

To address the challenges of bimanual tasks, we present a method for predicting bimanual actions, following the framework established by  \cite{shridhar2022}. Our model takes as input a 3D voxel grid, proprioception data, and a language goal. 
The voxel grid is assembled by combining sensory data streams from several RGB-D cameras. A PerceiverIO transformer learns per-voxel and language features. For each robot arm, the output is a discretized action consisting of a 6-DoF end-effector pose, a gripper state and an additional flag for collision-aware motion planning.
We choose \peract as the basis for our application because the voxel-based representation makes it robust to changes in the view pose.
Unlike \rvt, which suffers from occlusion issues due to its reliance on rendered virtual images, \peract, directly handles raw input data, avoiding these problems.
Additionally, unlike \alohaact, which relies on joint angles and may struggle with adaptability due to the need for similar demonstrations in the joint space, \peract and \rvt are robot-agnostic and can be transferred to other robots with different degrees of freedom.
Dealing with the question to get bimanual control, a naive approach to bimanual control would be to instantiate two independent instances of \peract as separate agents, each of them controlling one robot arm. We will refer to this as two \textit{independent} agents. Coordination between these two agents is only possible with visual perception.
Another drawback is that the voxel representation is stored twice, resulting in an increased memory usage.  
Another approach, is to adopt a \leaderfollower based architecture.
Here, the predicted output from one agent is passed to the second network. Hence, the prediction of the second network is based on the prediction of the first network. Once both action predictions are inferred, they are executed simultaneously.
While this offers the advantage of communication, this approach yet suffers from large memory consumption and fixed roles.
\\
To overcome the drawback of the voxel representation and facilitate communication without fixed roles, we propose a novel transformer module. To achieve this, we split up the latent space between the arms and implemented a combined self-attention.
Hence, our approach utilizes a single Perceiver IO as backbone to predict both actions simultaneously. Besides sharing the voxel representation, another advantage is that the proprioception of the two agents can be shared.

\subsection{Expert Demonstrations}

Demonstrations consist of a set of action tuples for each robot arm that are executed simultaneously. 
Following previous works \cite{shridhar2022}, we assume a dataset  $\mathcal{D} = \{\zeta_1, \zeta_2, \ldots, \zeta_n\}$ of $n$ expert demonstrations, each paired with language goals $\mathcal{G} = \{\mathbf{l}_{1}, \mathbf{l}_{2}, \ldots, \mathbf{l}_{n}\}$. 
For the bimanual setup we assume each demonstration contains two actions. Thus, each demonstration $\zeta_{i}$ is a sequence of
continuous actions $\mathcal{A} = \{(a^{r}_1, a^{l}_1), (a^{r}_2, a^{l}_2), ...., (a^{r}_t, a^{l}_t)\}$, where the superscripts $r$ and $l$ denote the right or the left  robot arm.
Each action $a$ contains the 6-DoF pose, gripper open state, and whether the motion-planner used collision avoidance to reach an intermediate pose: $a = \{a_{\textrm{pose}}, a_{\textrm{open}}, a_{\textrm{collide}}\}$. Additionally, we capture visual observations for $\mathcal{O} = \{\Tilde{o}_{1}, \Tilde{o}_{2}, \ldots \Tilde{o}_{t}\}$.
An observation $\Tilde{o}$ consists of RGB-D images from any number of cameras. For our simulated experiments, we use a total of five cameras and thus $\Tilde{o}_{\textrm{sim}} = \{o_{\textrm{front}}, o_{\textrm{left}}, o_{\textrm{right}}, o_{\textrm{wrist left}}, o_{\textrm{wrist right}} \}$. 
Each demonstration $\zeta$ is a sequence of continuous actions $\mathcal{A}$ paired with observations $\mathcal{O}$.

\subsection{Keyframe Extraction}

During the demonstrations salient keyframes are identified among the recorded visual and proprioceptive sensor data, which are used for training.
Similar to prior work \cite{shridhar2022} and \cite{james2022coarseexpansion} for any given 
demonstration $\zeta_i$ of length $m$ we identify keyframes $ k_1 < k_2 < .. k_n $.
We extend the heuristic to the bimanual case and define a keyframe if 
\begin{enumerate}[topsep=0pt, nosep]
    \item The gripper state of one of the robot has changed, or
    \item A robot reached the end of its executed trajectory, \ie the pose of an end-effector is no longer changing.
\end{enumerate}
The discretisation of keyframe actions $\mathbf{k}$  facilitates the conceptualization of our BC agent's training paradigm as a classification task, specifically focusing on the identification of the 'next best action'.

\subsection{Action Inference}

Our goal is to learn bimanual action-centric representations~\cite{gibson2014ecological} and retrieve a 6-DoF pose from a voxel for each arm. Hence the 6-DoF pose is split into a translation, a rotation and a gripper state. 
To this end, we utilize a 3-D voxel grid~\cite{moravec1996robot,30724} to represent both the observation and action space. The advantage is that such a representation is view-point independent compared to \alohaact.
The voxel grid $\mathbf{v}$ is reconstructed from several RGB-D images $\Tilde{o}$ and fused through triangulation $\Tilde{o} \Rightarrow \mathbf{v}$ from known camera extrinsics and intrinsics.
To allow for a fine-grained action representation we use $100^3$ voxels with size of $\SI{0.01}{\meter}$ to cover a workspace area of $\SI{1.0}{\meter}^3$.
\\
The translation is identified as the voxel nearest to the gripper fingers' center. Rotation is quantified into discrete \SI{5}{\degree} intervals for each rotational axis. The state of the gripper $a_{open}$, either open or closed, is represented through a binary value. Similarly, the 'collide' parameter $a_{collide}$ is binary, and indicates if the motion-planner should avoid the voxel grid. This binary mechanism is pivotal for enabling tasks that require both contact-based actions, like opening a drawer, and non-contact maneuvers, such as reaching a handle without physical contact.

\subsection{Training}

We extend the loss function as in \cite{shridhar2022} to the bimanual setup and thus the loss function results in 
\begin{equation*}
    \mathcal{L}_{\textrm{total}} = \sum_{i \in \mathcal{X}} \mathcal{L}^{\textrm{right}}_{\textrm{i}} + \sum_{i \in \mathcal{X}} \mathcal{L}^{\textrm{left}}_{\textrm{i}}
\end{equation*}
with $\mathcal{X} \in \{\textrm{trans}, \textrm{rot}, \textrm{open}, \textrm{collide}\}$ and $\mathcal{L}_{\textrm{i}} = \mathbb{E}_{Y_{\textrm{i}}}[\textrm{log} \mathcal{V}_{\textrm{i}}]$, where
\[
\begin{array}{ll}
\mathcal{V}_{\textrm{trans}} = \textrm{softmax}(\mathcal{Q}_{\textrm{trans}}((x,y,z) | \mathbf{v}, \mathbf{l})) & \quad \mathcal{V}_{\textrm{rot}} = \textrm{softmax}(\mathcal{Q}_{\textrm{rot}}((\psi, \theta, \phi) | \mathbf{v}, \mathbf{l})) \\[1em]
\mathcal{V}_{\textrm{open}} = \textrm{softmax}(\mathcal{Q}_{\textrm{open}}(\omega | \mathbf{v}, \mathbf{l})) & \quad \mathcal{V}_{\textrm{collide}} = \textrm{softmax}(\mathcal{Q}_{\textrm{collide}}(\kappa | \mathbf{v}, \mathbf{l}))
\end{array}
\]
Similar to \peract, we also augment $\mathbf{v}$ and $\mathbf{k}$ with translation and rotation perturbations for robustness, keeping other parameters, such as the optimizer, the same.

\section{Evaluation}
\label{sec:evaluation}

We study the efficacy of robotic bimanual manipulation. To this end, we compare our method against the following baselines.
\begin{enumerate*}[label=\alph*.)]
    \item \alohaact: A transformer network with action chunking that outputs joint positions from camera inputs.
    \item \rvtlf: Two Robotic View Transformer (RVT) as a \leaderfollower architecture.
    \item \peractlf: Two Perceiver Actor networks as a \leaderfollower architecture.
    \item \peracttwo: A single bimanual Perceiver Actor network as described in \autoref{sec:method}.
\end{enumerate*}

The \leaderfollower architecture consists of two networks, where the output of one network is fed as a prediction to the other, and then both actions are executed. For \peract we updated the
network architecture as well as reduced floating point precision\footnote{https://github.com/ishikasingh/YARR/commit/875f636} 
resulting in a significant reduction of the training time.
We report on the task success rate as well as on the training time.
For simulated experiments, we use two Franka Panda robots with parallel grippers.
\ifanonymized
To demonstrate that our method is robotic-agnostic, we also test with a humanoid in real world.
\else
To demonstrate that our method is robotic agnostic, we also test with the humanoid robot ARMAR-6 \cite{Asfour.2019} in real world.
\fi
For \alohaact we set 
$\Tilde{o}_{\textrm{sim}} = \{o_{\textrm{front}}, o_{\textrm{wrist left}}, o_{\textrm{wrist right}} \}$ to minimize the network input.

\subsection{Simulation}

{\setlength{\tabcolsep}{0.4em}
\begin{table}[h!]
\centering
\begin{tabular}{lrrrrrrrr}
\toprule
 \multirow{2}{*}{Method} & \multicolumn{7}{c}{ Task success \arrowup } \\
\cmidrule(l){2-8}
 & {(a) box} & {(b) ball} & {(c) buttons} & {(d) plate} &  {(e) drawer} & {(f) fridge} & {(g) handover} \\
\midrule
\alohaact      & \SI{0}{\percent}  & \SI{36}{\percent} & \SI{4}{\percent}  & \SI{0}{\percent}  & \SI{13}{\percent}  & \SI{0}{\percent}  & \SI{0}{\percent}  \\
\rvtlf        & \SI{52}{\percent} & \SI{17}{\percent} & \SI{39}{\percent} & \SI{3}{\percent}  & \SI{10}{\percent} & \SI{0}{\percent}  & \SI{0}{\percent}  \\
\peractlf     & \starstar \SI{57}{\percent} & \SI{40}{\percent} & \SI{10}{\percent} & \SI{2}{\percent}  & \starstar  \SI{27}{\percent} & \SI{0}{\percent}  & \SI{0}{\percent}  \\
\peracttwo     &  \SI{6}{\percent} & \starstar  \SI{50}{\percent} & \starstar \SI{47}{\percent} & \starstar  \SI{4}{\percent}  & \SI{10}{\percent} & \starstar  \SI{3}{\percent}  & \starstar  \SI{11}{\percent}  \\
\cmidrule(l){2-8}
 & {(h) laptop} & {(i) rope} & {(j) dust} & {(k) tray} & {\makecell{(l) handover \\ easy}} & {(m) oven} \\ \cmidrule(l){2-8}
\alohaact      & \SI{0}{\percent}  & \SI{16}{\percent}  & \SI{0}{\percent}  & \SI{6}{\percent}  & \SI{0}{\percent}   & \SI{2}{\percent}  \\
\rvtlf      & \SI{3}{\percent}  & \SI{3}{\percent}  & \SI{0}{\percent}  & \SI{6}{\percent} & \SI{0}{\percent}    & \SI{3}{\percent}  \\
\peractlf   & \SI{11}{\percent} & \SI{21}{\percent}  & \starstar  \SI{28}{\percent} & \starstar \SI{14}{\percent} & \SI{9}{\percent}  & \SI{8}{\percent} \\
\peracttwo & \starstar \SI{12}{\percent}  & \starstar \SI{24}{\percent} & \SI{0}{\percent}  & \SI{1}{\percent}  & \starstar \SI{41}{\percent}   & \starstar  \SI{9}{\percent}  \\

\bottomrule
\end{tabular}
\vspace{0.5\normalbaselineskip}
\caption{Performance of different methods on various tasks.}
\label{tab:evaluation_results_single}
\vspace{-0.5\baselineskip}
\end{table}}

\begin{wraptable}{r}{0.44\textwidth}
    \vspace{-\normalbaselineskip}
    \centering
\begin{tabular}{@{}lrr@{}}
\toprule

\multicolumn{1}{c}{Method}                        & \multicolumn{1}{c}{\makecell{avg. task \\ success \arrowup}} & \multicolumn{1}{c}{\makecell{avg. training \\ time \arrowdown}}  \\ \midrule
\alohaact                                   & \SI{5.9}{\percent}                                           & \SI{80}{\hour}                                                                           \\
\rvtlf                                      & \SI{10.5}{\percent}                                           & \SI{231}{\hour}                                                          \\ 
\peractlf                                   & \starstar \SI{17.5}{\percent}                                          & \SI{89}{\hour}                                                                            \\
\peracttwo                                  & \SI{16.8}{\percent}                                          & \starstar  \SI{54}{\hour}                                                                             \\
\bottomrule
\end{tabular}
\caption{Overview of the average task success rate and average training time with respect to different input image size for 100 demonstrations.}
\label{tab:evaluation_overview}
\vspace{-\normalbaselineskip}
\end{wraptable}

\vspace{-0.5\normalbaselineskip}

We conduct our primary experiments in simulation for reproducibility and benchmarking. The environment is similar to \cite{shridhar2022}. RGB-D sensors are positioned at the front, left shoulder, right shoulder, and on the wrist. %
All cameras are noiseless and have a resolution of $256 \times 256$. 
The increase in image resolution ensures future comparability with other methods that require it.

We trained all tasks individually, as this allows for a more refined analysis and different coordination types can be distinguished. We also note that while it is possible to train multi-task agents, not all methods accommodate this setting. We used 100 demonstrations for each task.
All single tasks were trained on a NVIDIA A40 GPUs for up to 100k iterations.
For each method, the batch size was maximized to fit into the GPU memory.
Every 10k-th checkpoint was evaluated with 100 episodes and the best checkpoint was finally evaluated on a separate test set.
\autoref{tab:evaluation_results_single} lists the task success rate for a single-task agents for each individual task.

The difference in the success rate of image-based methods, such as \alohaact and \rvt can be explained by the symmetry in the tasks. For example the both robot arms are the exact same model making it difficult to distinguish between them. Furthermore, other challenges for \alohaact include that the demonstrations can have high variations due to the randomization of the spawned objects or the generated motion planning paths. Both aspects are challenging because \alohaact predicts joint angles and not a 6-DoF pose.

\subsection{Discussion of Failure Types}

In the following, we will briefly discuss common failures.
During the handover task, a robot may experience a collision with another robot due to inadequate spatial awareness or timing errors. Grasping failures are also common, often due to misalignment of the gripper, leading to an inability to securely grasp the object.  For the picking up the plate or the tray tasks, a robot arm may miss the object entirely. This could be a result of insufficient demonstrations or errors in motion planning. Lastly, inserting an item into a drawer introduces complexities such as the requirement for a temporal dependencies: A robot may fail to open the drawer.
These failures underscore the challenges in bimanual robotic manipulation when interacting with the scene and the need for more sophisticated motion planning strategies.

\subsection{Real-World}

\begin{figure}[pbt]
    \centering
    \captionsetup[subfigure]{justification=centering}
    
    \begin{subfigure}[b]{.245\textwidth}
        \centering
        \includegraphics[width=\textwidth]{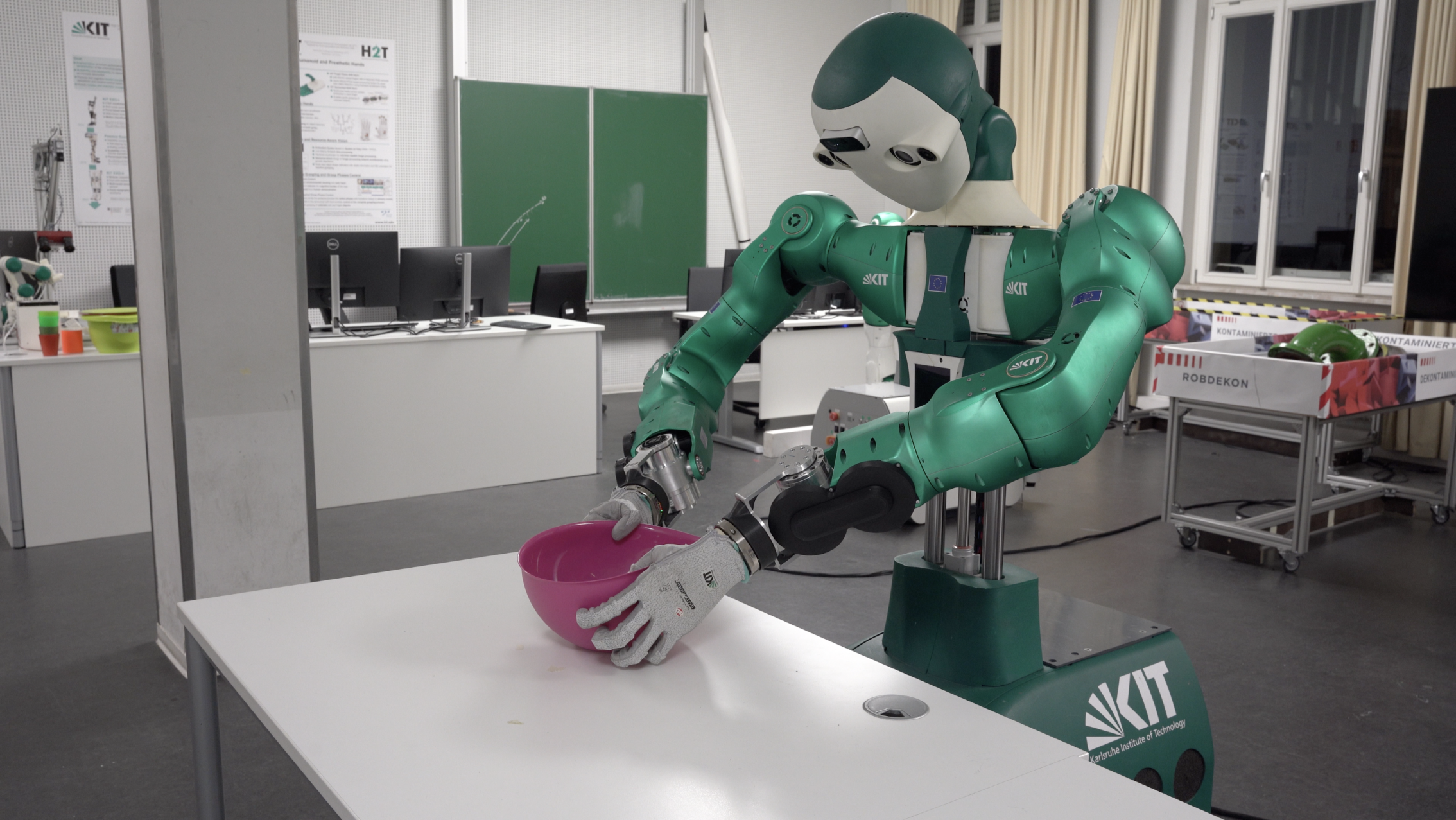}
        \caption{lifting a bowl}
    \end{subfigure}%
    \begin{subfigure}[b]{.245\textwidth}
        \centering
        \includegraphics[width=\textwidth]{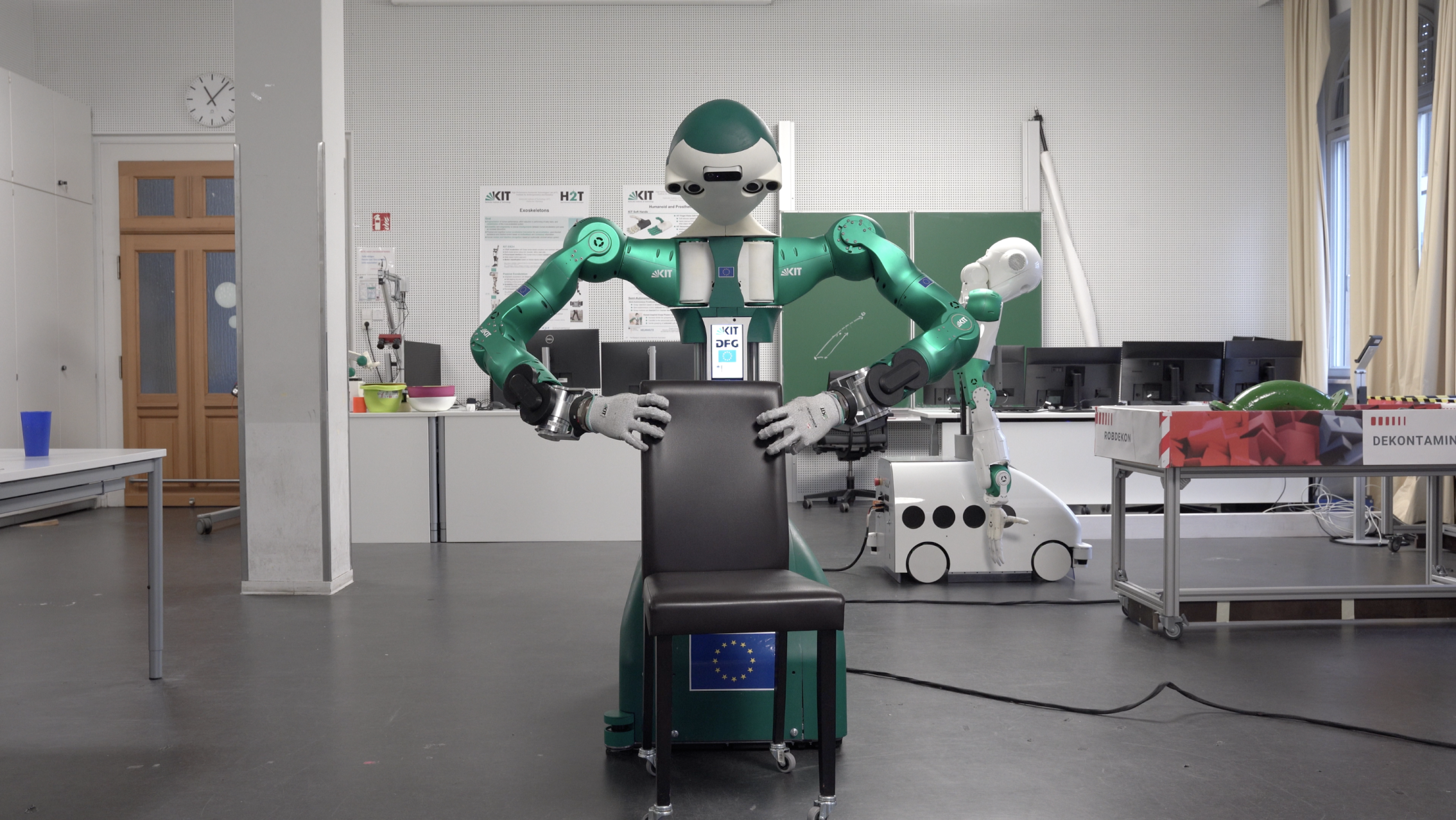}
        \caption{pushing a chair}
    \end{subfigure}%
    \begin{subfigure}[b]{.245\textwidth}
        \centering
        \includegraphics[width=\textwidth]{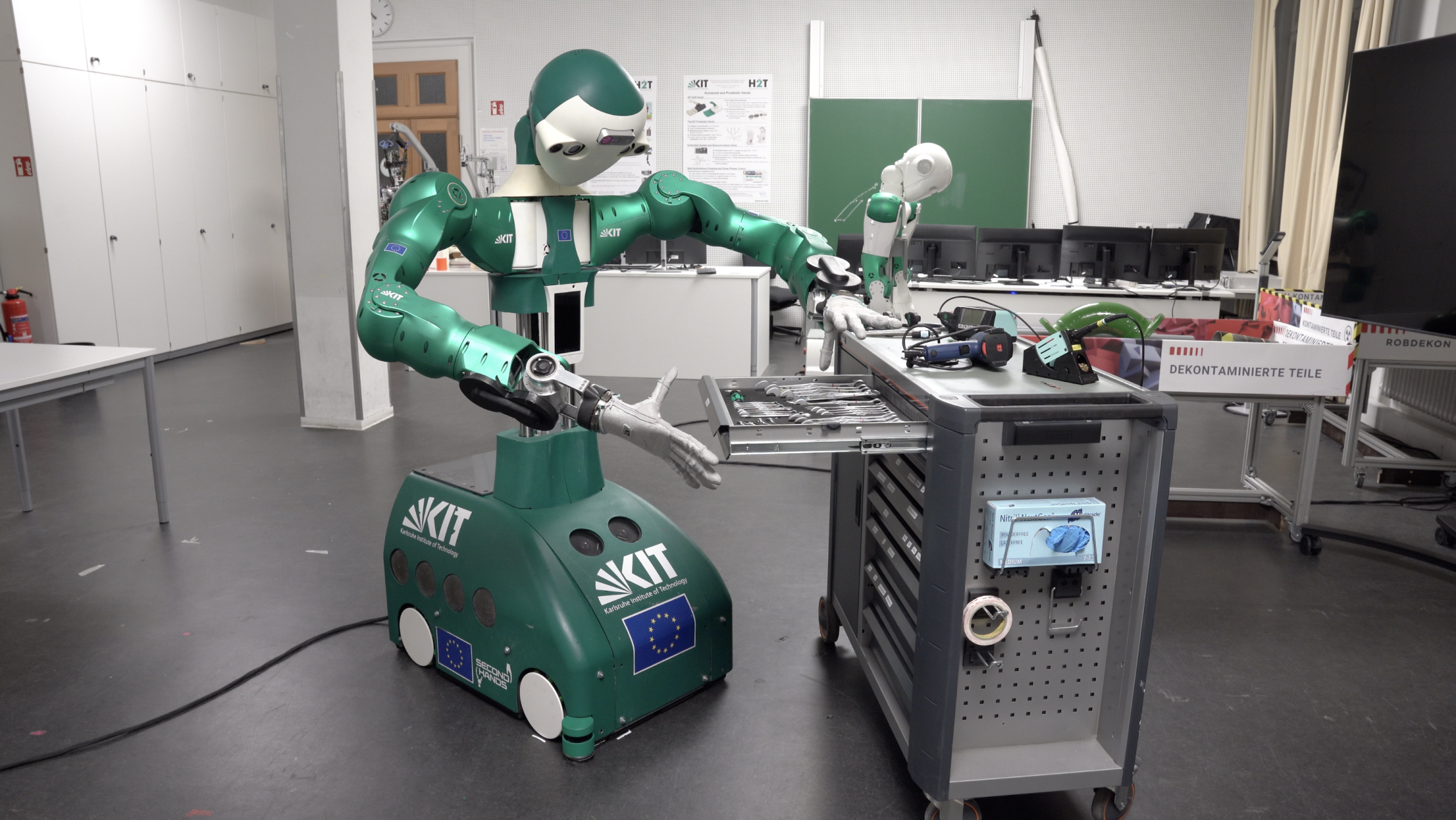}
        \caption{storing away a tool}
    \end{subfigure}%
    \begin{subfigure}[b]{.245\textwidth}
        \centering
        \includegraphics[width=\textwidth]{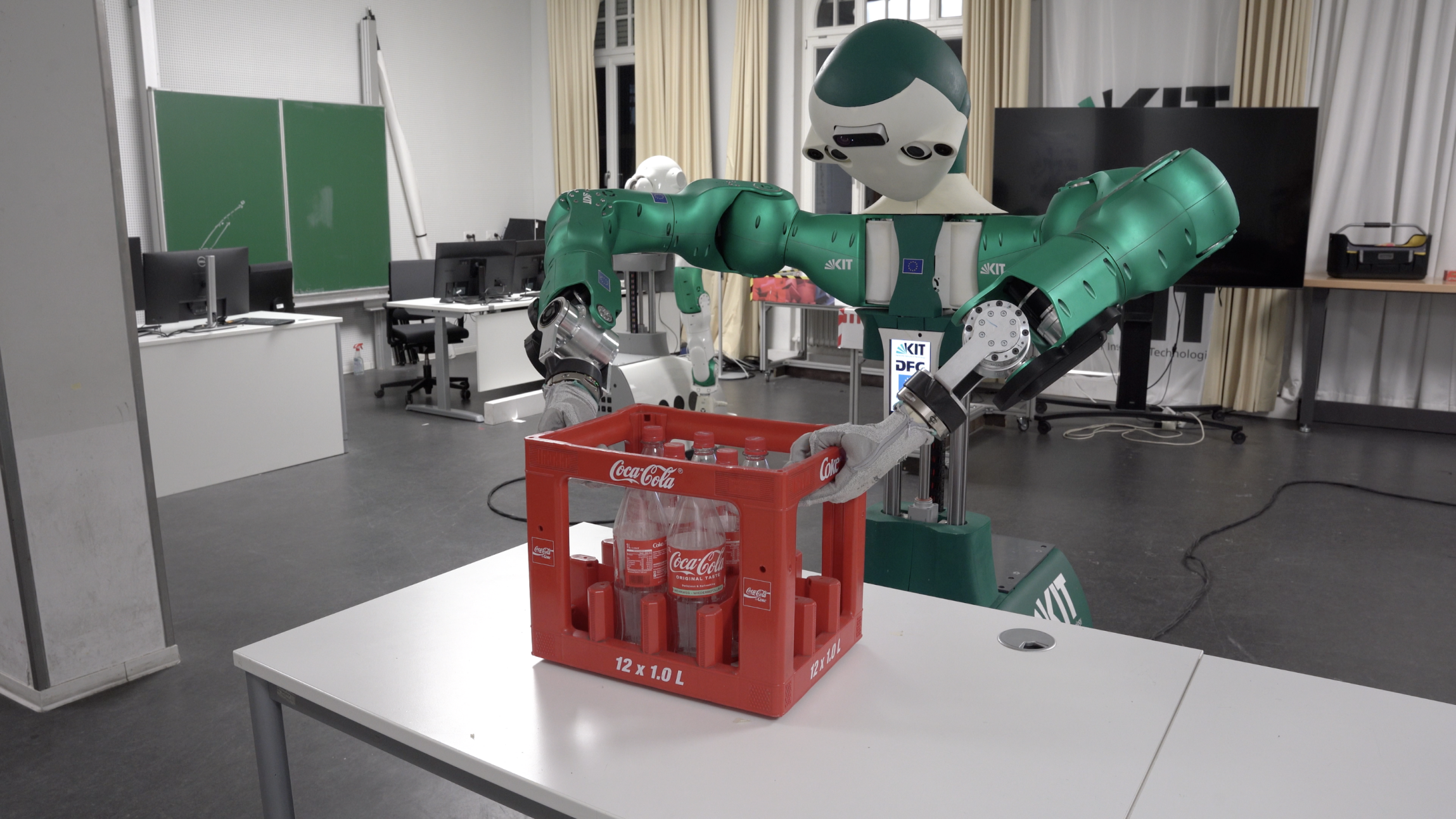}
        \caption{getting a coke}
    \end{subfigure}

    \vspace{0.5\baselineskip}

     \caption{Selected snapshots of the real world experiments showing different tasks. A video showing the full experiments is available on the project's to website.}
     \label{fig:experiments_real_world}
\end{figure}

\ifanonymized
To also demonstrated that the framework is robot-agnostic, the method has been integrated into a humanoid robot.
\else
To also demonstrated that the framework is robot-agnostic, the method has been integrated into the humanoid robot ARMAR-6 \cite{Asfour.2019, Peller.2023}.
\fi
The robot is equipped with an Azure Kinect RGB-D sensor with a resolution of $1920 \times 1080 $ and thus for the experiments only a single camera is used, \ie $\Tilde{o}_{\textrm{real}} = \{o_{\textrm{front}} \}$.  
The voxel size is set to $50 \times 50 \times 50$ to speed up training, but reduces accuracy.
A Cartesian waypoint controller
\ifanonymized
\else
 as used in \cite{Grimm.2021}
\fi
moves the end-effector to the predicted targets. For each task a single demonstration $\zeta_i$ is recorded using Kinesthetic teaching instead of VR.
Overall, four different tasks have been demonstrated to the robot.
Three of the tasks require a synchronous coordination of both arms, such as \textit{``lifting a bowl''} or \textit{``pushing a chair''}. The fourth task, \textit{``put away the tool''}, requires spatial coordination.
While it is possible to quantify results in real world, reproducing them is challenging due to a lack of hardware as well as to object poses and sensor noise.
\autoref{fig:experiments_real_world} shows the tasks for the real-world experiments. A video showing the experiments is available on the website.

\section{Conclusion}
\label{sec:conclusion}

In this work, we presented a robotic manipulation benchmark specifically designed for bimanual robotic manipulation tasks. We open-source 13 new tasks with 23 unique task variations, each requiring a high degree of coordination and adaptability.
We extended two existing methods to bimanual manipulation and run them with another method on the benchmark.
Additionally, we presented \peracttwo -- a perceiver-actor agent for bimanual manipulation. Due to the architecture design the method can easily be transferred to other robots, such as a humanoid, as the policy outputs a 6-DoF pose and control is separated.
Our investigation reveals that our method is the most successful in 9 out of 13 tasks and performs effectively in real-world settings while also having the fastest training time. For the average task success rate, both \peractlf and \peracttwo outperformed image-based methods, achieving average task success rates of \SI{17.5}{\percent} and \SI{16.8}{\percent}, respectively.

\paragraph{Limitations and Future Work}
None of the methods is able to achieve a sufficiently high success rate, which can be explained by the complexity of bimanual manipulation.
Another limiting factor is that methods using discretized actions rely on a sampling-based motion planner for successful execution.
We acknowledge that tasks in the benchmarks are challenging, and we are looking forward for the community to pick up on this challenge.
Future work will focus on extending the benchmark by adding more state-of-the-art methods and including mobile manipulation.

\clearpage
\acknowledgments{This work has been supported by 
Amazon under the research project "Robotic Manipulation in Densely Packed Containers".
We would further like to thank  Yi Li, Bernie Zhu, Jonas Frey, Aaron Walsman for the insightful suggestions and providing feedback. Moreover, Christian Dreher, Felix Hundhausen and Timo Birr for making the real-world robot experiments possible.}

\bibliography{references}  %

\clearpage

\appendix

\section{Description of Tasks}
\label{sec:description_of_tasks}

To model the complexity of a task we report in \autoref{tab:task_details} on the average time length of the demonstrations, the number of identified keyframes, number of items and the number of task variations
Keyframes are discrete frames in a continuous stream of data, and the number of keyframes is a measure of the number of actions necessary to complete a task.

\begin{table}[tp]
\centering
\begin{tabular}{@{}lcccc@{}}
\toprule
\multicolumn{1}{c}{Task}                     & duration & \# keyframes & \# items & \# variations \\ \midrule
(a) push box             & \SI{4.33}{\second}     & 2.1         & 1       & 1            \\
(b) lift a ball          & \SI{4.40}{\second}     & 4.0         & 1       & 1            \\
(c) push two buttons     & \SI{3.47}{\second}     & 4.0         & 3       & 5           \\
(d) pick up a plate      & \SI{6.47}{\second}     & 6.6         & 1       & 1            \\
(e) put item in drawer   & \SI{5.57}{\second}     & 8.4         & 5       & 3            \\
(f) put bottle in fridge & \SI{9.70}{\second}     & 7.8         & 2       & 1            \\
(g) handover an item     & \SI{7.63}{\second}     & 7.6         & 5       & 5            \\
(h) pick up notebook     & \SI{3.97}{\second}     & 7.2         & 1       & 1            \\
(i) straighten rope      & \SI{3.83}{\second}     & 5.9         & 1       & 1            \\
(j) sweep dust pan       & \SI{4.93}{\second}     & 7.3         & 1       & 1            \\
(k) lift tray            & \SI{3.77}{\second}     & 5.1         & 1       & 1            \\
(l) handover item (easy) & \SI{7.17}{\second}     & 7.5         & 1       & 1            \\ 
(m) take tray out of oven & \SI{10.13}{\second}    & 8.7         & 2       & 1            \\
\bottomrule
\end{tabular}%
\vspace{0.5\normalbaselineskip}
\caption{Properties of the bimanual tasks. We report on the average length of the task demonstration in seconds. The average number of extracted keyframes of the task, the number of items that the robot can interact with and the task variations.}
\label{tab:task_details}
\end{table}

\subsection*{(a) push box}

\begin{wrapfigure}{l}{0.4\textwidth}
    \centering
    \vspace{-\normalbaselineskip}
    \includegraphics[width=.38\textwidth,keepaspectratio=true]{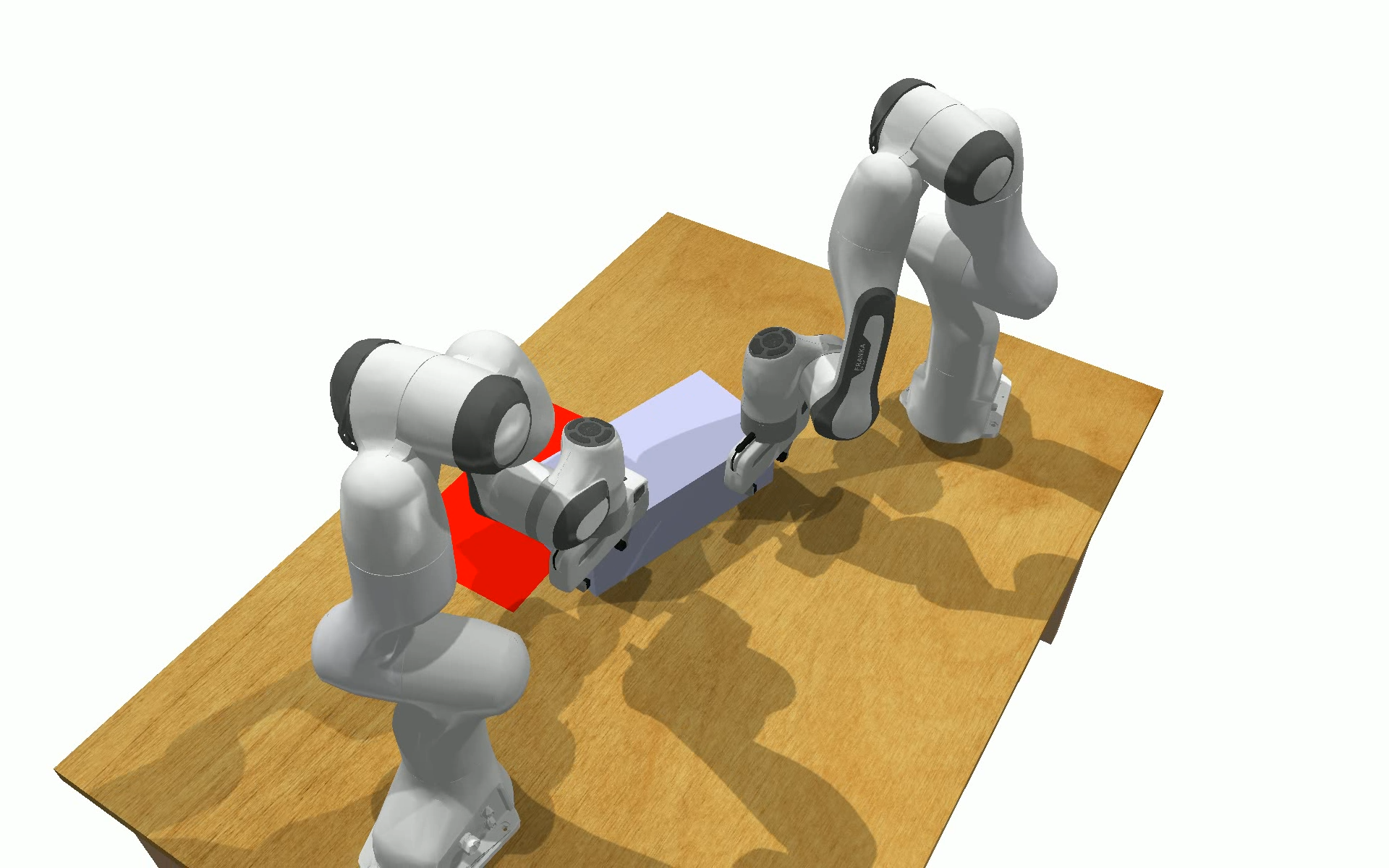}
    \vspace{-\normalbaselineskip}
\end{wrapfigure}

\textbf{Task Description:} The robot's task is to push a heavy box using both arms to move it to a designated target area.

\textbf{Success Metric:} The task is considered successfully completed when the box reaches the targetarea.

\textbf{Objects:} The task involves a large box and a target area.

\textbf{Coordination Challenges:} 
The primary challenge lies in the weight of the box, which is set to \SI{50}{\kilo\gram}, making it very difficult for a single arm to push.

\textbf{NB:} This task cannot be solved with one robot due to the weight of the box.

\textbf{Language Instructions:} \textit{Push the box to the red area.}

\subsection*{(b) lift a ball}

\begin{wrapfigure}{l}{0.4\textwidth}
    \centering
    \vspace{-\normalbaselineskip}
    \includegraphics[width=.38\textwidth,keepaspectratio=true]{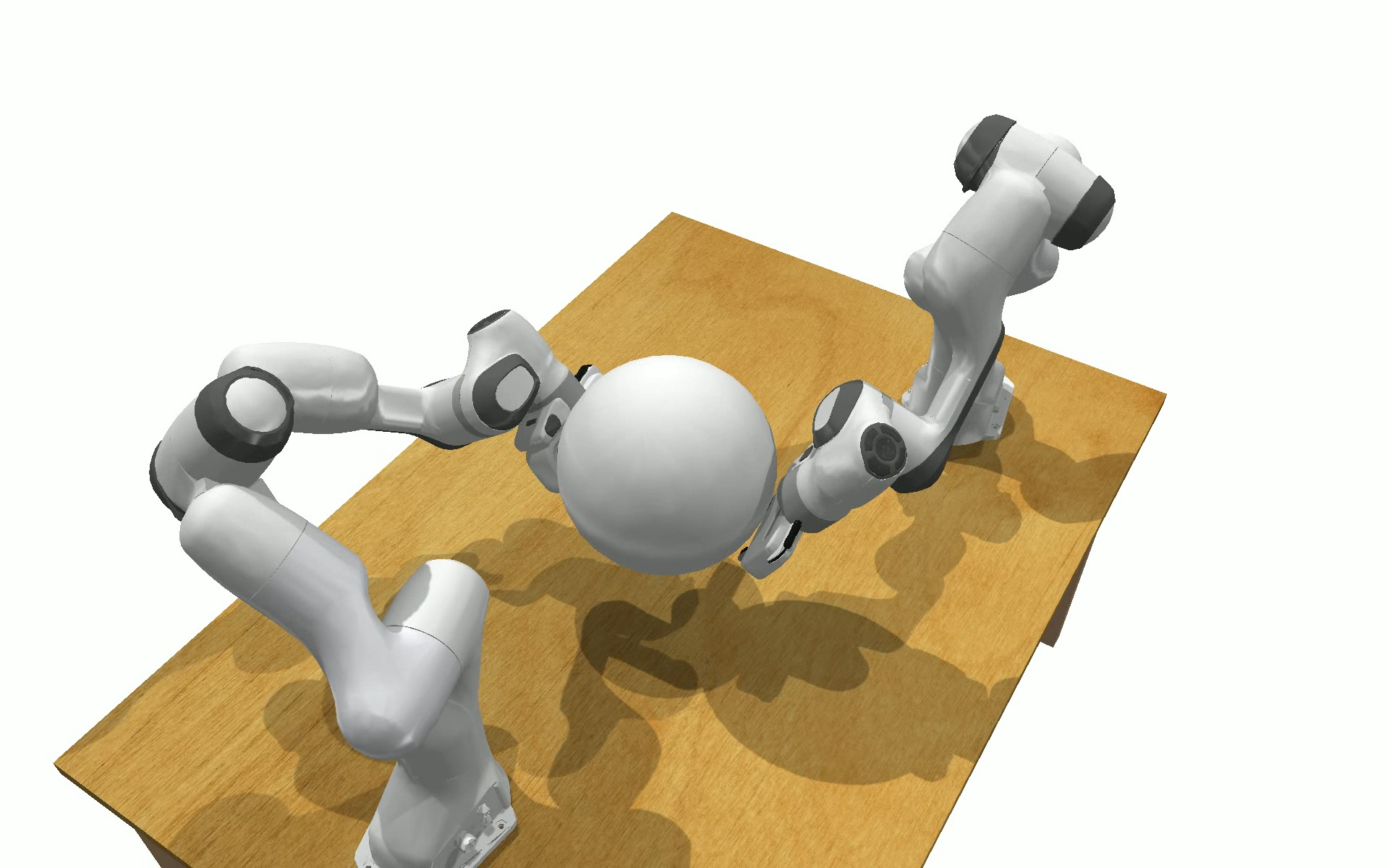}
    \vspace{-\normalbaselineskip}
\end{wrapfigure}

\textbf{Task Description:} The robot's task is to grasp and lift a large ball using both arms.

\textbf{Success Metric:} The task is considered successfully completed when the ball is lifted to a height above \SI{0.95}{\meter}.

\textbf{Objects:} The task involves a large ball.

\textbf{Coordination Challenges:} The primary challenge involves coordinated non-prehensile manipulation, as the ball cannot be grasped by the gripper. This requires careful coordination during the lifting motion.

\textbf{NB:} This task is impossible to solve with one robot due to the size of the object.

\textbf{Language Instructions:} \textit{Lift the ball.}

\subsection*{(c) push two buttons}

\begin{wrapfigure}{l}{0.4\textwidth}
    \centering
    \vspace{-\normalbaselineskip}
    \includegraphics[width=.38\textwidth,keepaspectratio=true]{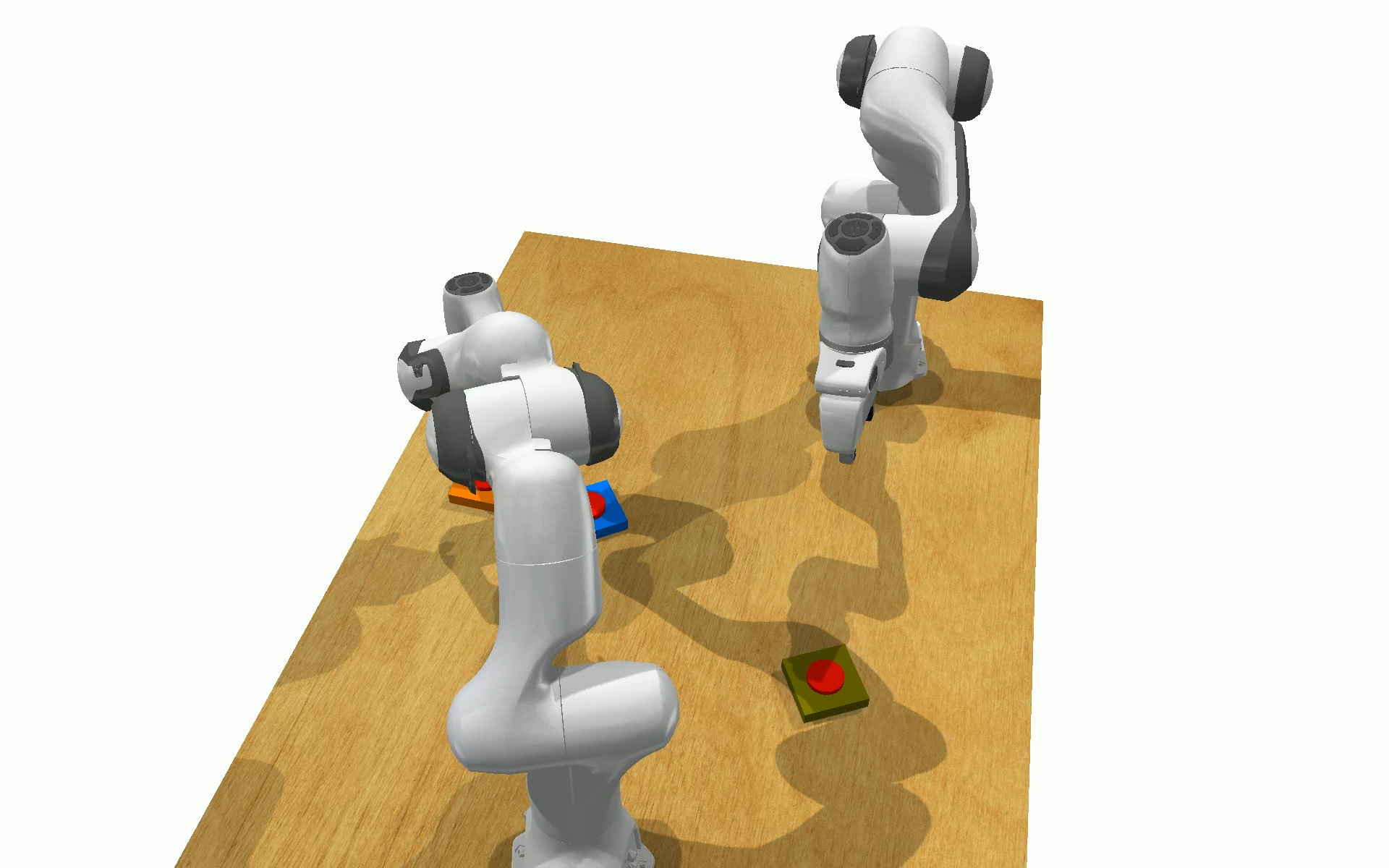}
    \vspace{-\normalbaselineskip}
\end{wrapfigure}

\textbf{Task Description:} The robot's task is to push two out of three buttons in an environment where the colors of the buttons are randomized. The goal is to press two specified buttons at the same time.

\textbf{Success Metric:} The task is considered successfully completed when both specified buttons are pressed simultaneously.

\textbf{Objects:} The task involves three buttons with different colors and a differently colored base.

\textbf{Coordination Challenges:} The primary challenge is the synchronous button press. The randomization of the button colors adds complexity compared to standard tasks.

\textbf{NB:} This task is impossible to solve with one robot as two buttons need to be pressed simultaneously.

\textbf{Language Instructions:} \textit{Push the (color A) and the (color B) button.}

\subsection*{(d) pick up a plate}

\begin{wrapfigure}{l}[0cm]{0.4\textwidth}
    \centering
    \vspace{-\normalbaselineskip}
    \includegraphics[width=.38\textwidth,keepaspectratio=true]{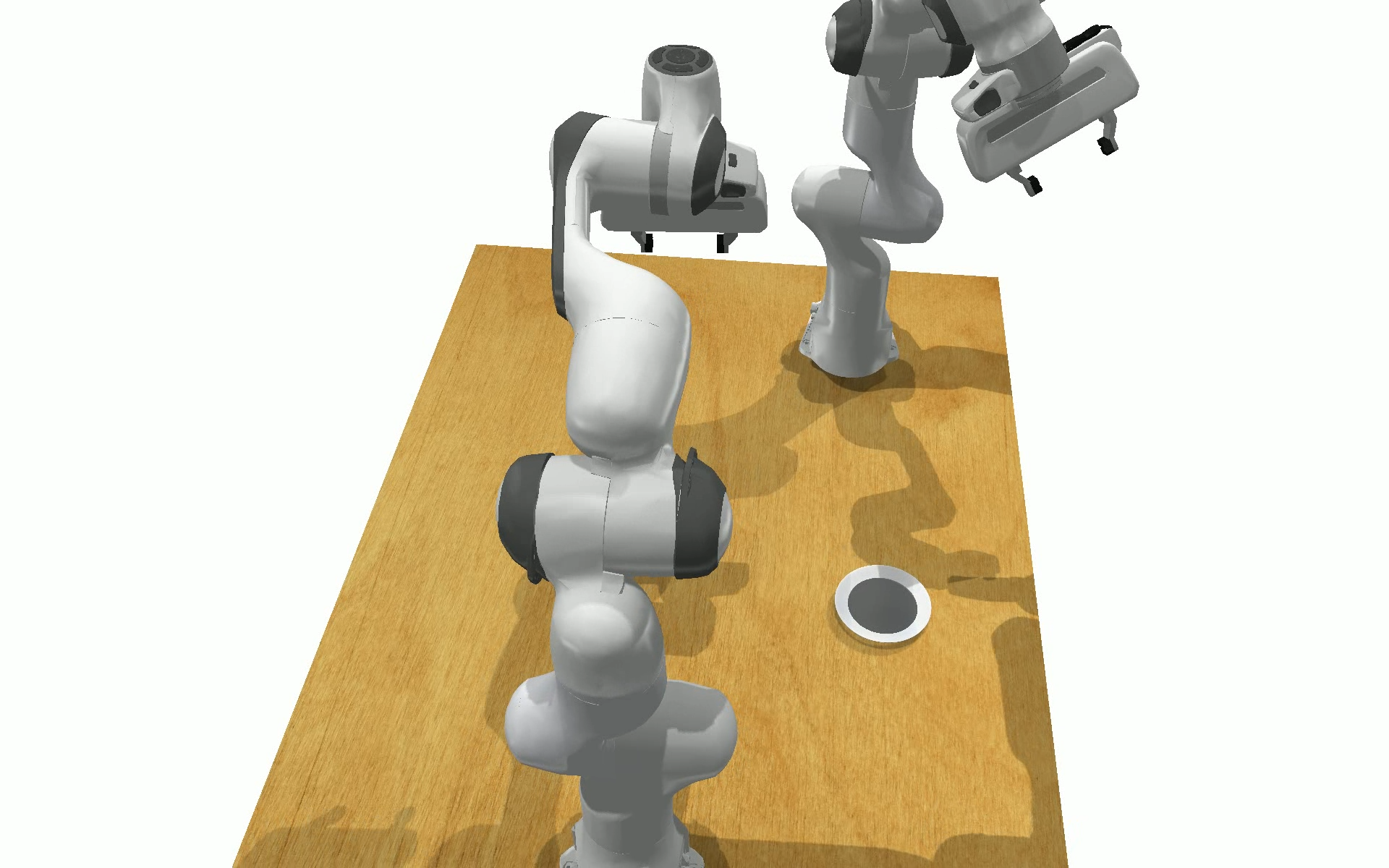}
    \vspace{-\normalbaselineskip}
\end{wrapfigure}

\textbf{Task Description:} The robot's task is to pick up a plate that is placed on a table. This involves grasping the plate and lifting it.

\textbf{Success Metric:} The task is considered successfully completed when the robot has securely grasped the plate and lifted it.

\textbf{Objects:} The task involves a single plate.

\textbf{Coordination Challenges:} The main challenges involve non-prehensile manipulation, as well as coordination during the lifting motion. The plate must be handled delicately to avoid slipping or tilting, which requires precise control.

\textbf{NB:} This task is difficult to solve with one robot because the plate's flat and smooth surface makes it hard to grasp securely with a single gripper. The coordination required to lift the plate without tilting or dropping it is challenging for one robot arm.

\textbf{Language Instructions:} \textit{Pick up the plate.}

\subsection*{(e) put item in drawer}

\begin{wrapfigure}{l}[0cm]{0.4\textwidth}
    \centering
    \vspace{-\normalbaselineskip}
    \includegraphics[width=.38\textwidth,keepaspectratio=true]{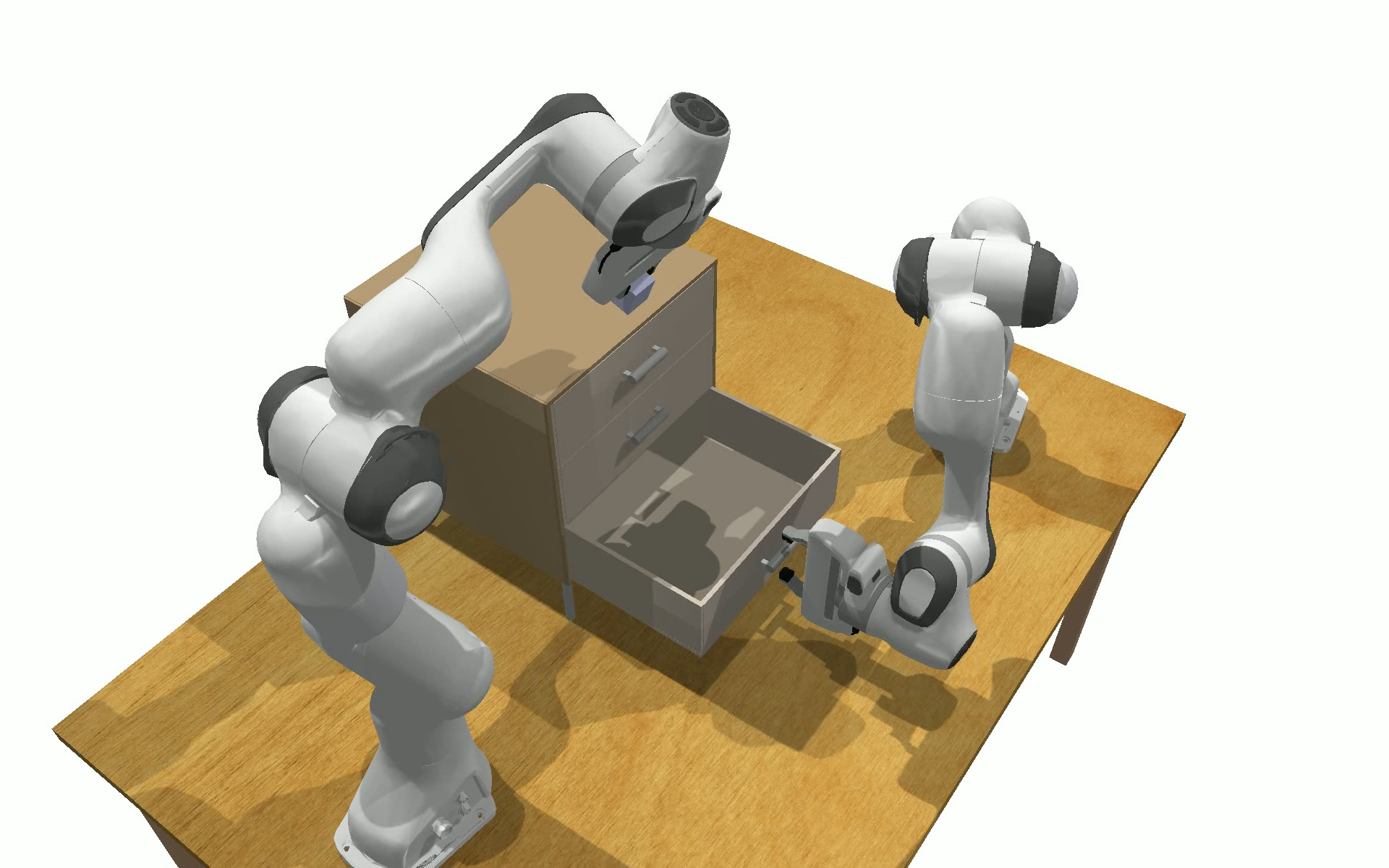}
    \vspace{-\normalbaselineskip}
\end{wrapfigure}

\textbf{Task Description:} The robot's task is to open a specific drawer in a cupboard and place an item into it. The correct drawer must be identified and opened before the item can be placed inside.

\textbf{Success Metric:} The task is considered successfully completed when the item is placed inside the specified drawer.

\textbf{Objects:} The task involves an item and a cupboard with three drawers.

\textbf{Coordination Challenges:} The primary challenge involves identifying the correct drawer and ensuring it is open before attempting to place the item inside. This requires coordination between the actions of opening the drawer and placing the item.

\textbf{Objects:} A cupboard with three drawers and an item on top

\textbf{Language Instructions:} \textit{Put the item into the (top, middle, bottom) drawer.}

\subsection*{(f) put bottle in fridge}

\begin{wrapfigure}{l}[0cm]{0.4\textwidth}
    \centering
    \vspace{-\normalbaselineskip}
    \includegraphics[width=.38\textwidth,keepaspectratio=true]{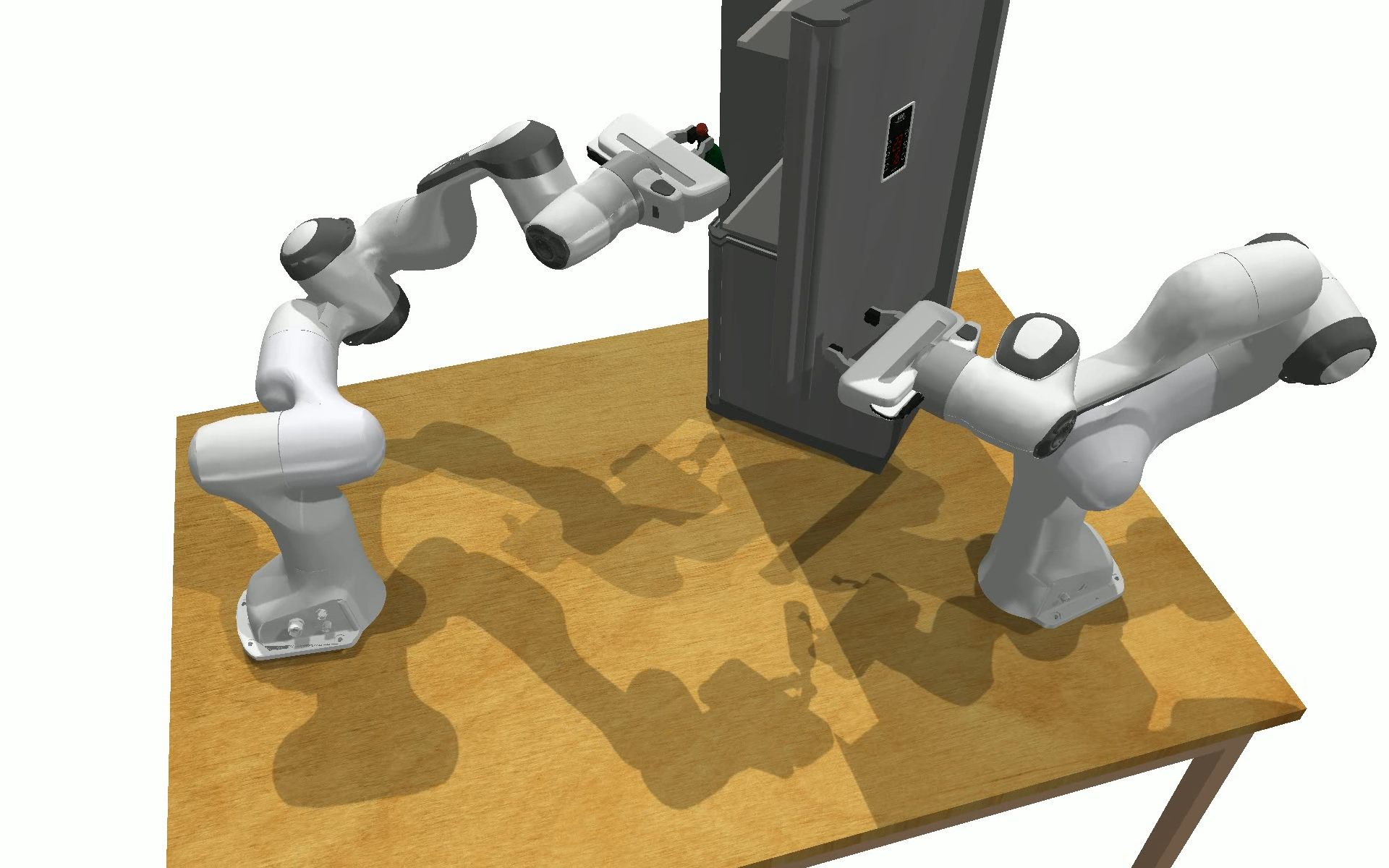}
    \vspace{-\normalbaselineskip}
\end{wrapfigure}

\textbf{Task Description:} The robot's task is to put a bottle into the fridge. This requires opening the fridge door, grasping the bottle, and placing it inside the fridge.

\textbf{Success Metric:} The task is considered successfully completed when the bottle is placed inside the fridge.

\textbf{Objects:} The task involves a bottle and a fridge.

\textbf{Coordination Challenges:} The primary challenges include:
- The fridge needs to be opened first.
- The bottle is difficult to grasp.
- Collision with the fridge needs to be avoided.
- Reachability is an issue as either the bottle or the fridge door can only be reached by one robot.

\textbf{Objects:} A bottle and a fridge

\textbf{NB:} This task requires two robots due to reachability issues.

\textbf{Language Instructions:} \textit{Put the bottle into the fridge.}

\subsection*{(g) handover an item}

\begin{wrapfigure}{l}[0cm]{0.4\textwidth}
    \centering
    \vspace{-\normalbaselineskip}
    \includegraphics[width=.38\textwidth,keepaspectratio=true]{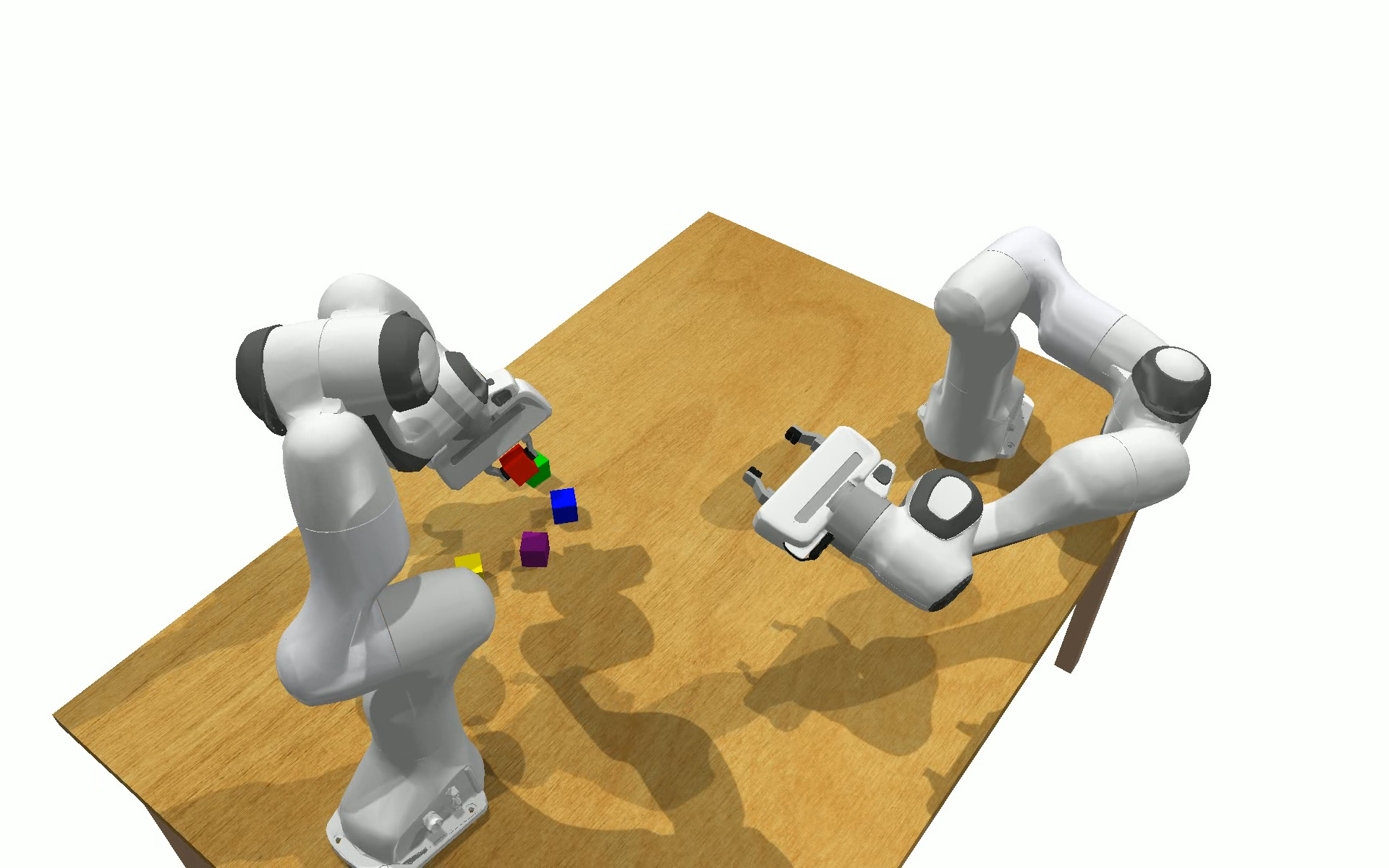}
    \vspace{-\normalbaselineskip}
\end{wrapfigure}

\textbf{Task Description:} The robot's task is to hand over the \textit{color} item. This involves identifying the correct item based on its color, grasping it, and lifting it to the required height.

\textbf{Success Metric:} The task is considered successfully completed when the robot has securely grasped the correct item and lifted it to a height of \SI{80}{\centi\meter}, while the other arm remains idle.

\textbf{Objects:} The task involves five items with different colors.

\textbf{Coordination Challenges:} The main challenge lies in correctly identifying the item based on its color as specified in the task description, and then coordinating the handover process.

\textbf{NB:} There are variations of this task: one with only three items instead of five, and a simpler task that involves only a block instead of colored cubes.

\textbf{Language Instructions:} \textit{Hand over the (red, green, blue, yellow) item.}

\subsection*{(h) pick up notebook}

\begin{wrapfigure}{l}[0cm]{0.4\textwidth}
    \centering
    \vspace{-\normalbaselineskip}
    \includegraphics[width=.38\textwidth,keepaspectratio=true]{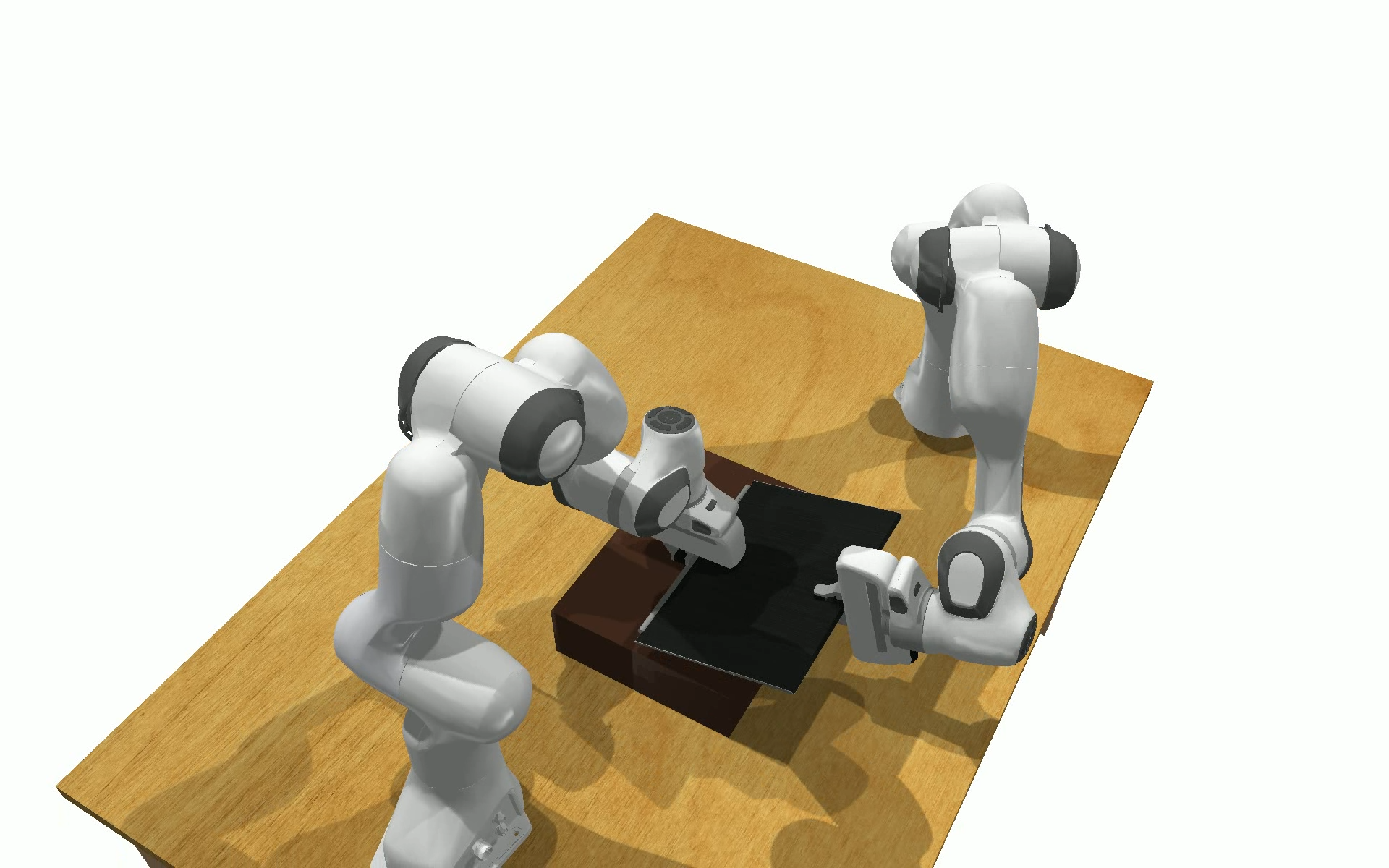}
    \vspace{-\normalbaselineskip}
\end{wrapfigure}

\textbf{Task Description:} The robot's task is to pick up a notebook that is placed on top of a block. This requires the robot to first manipulate the notebook into a position where it can be grasped.

\textbf{Success Metric:} The task is considered successfully completed when the robot has securely grasped the notebook and lifted it off the block.

\textbf{Coordination Challenges:} Since the notebook is resting on a block, the robot must perform non-prehensile manipulation, such as pushing or sliding, to reposition the notebook into a graspable orientation.

\textbf{Objects:} The task involves two primary objects: a notebook and a block.

\textbf{NB:} This task can be accomplished with a single robotic arm, though coordination is crucial for successful manipulation.

\textbf{Language Instructions:} \textit{pick up the notebook}

\subsection*{(i) straighten rope}

\begin{wrapfigure}{l}[0cm]{0.4\textwidth}
    \centering
    \vspace{-\normalbaselineskip}
    \includegraphics[width=.38\textwidth,keepaspectratio=true]{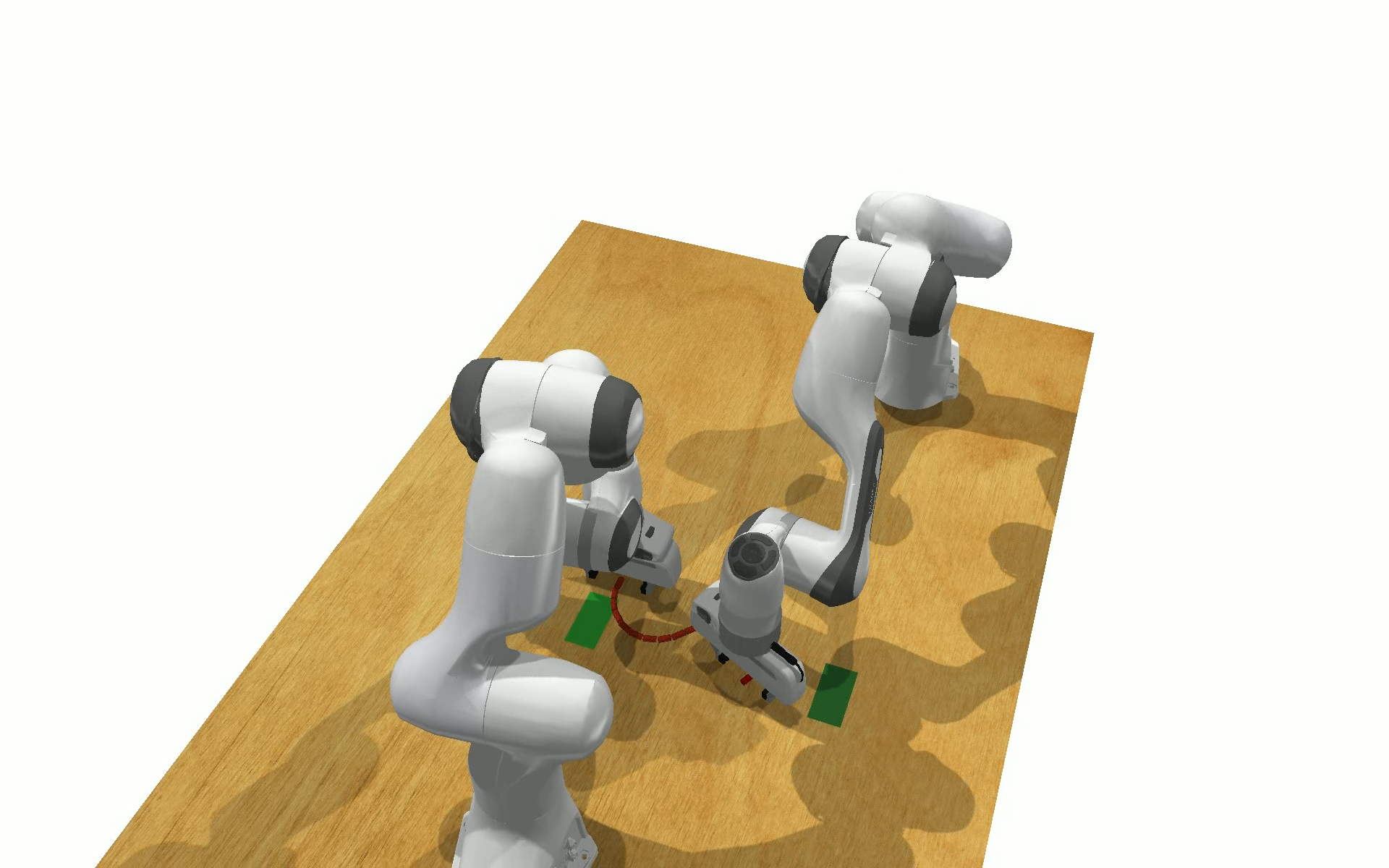}
    \vspace{-\normalbaselineskip}
\end{wrapfigure}

\textbf{Task Description:} The robot's task is to straighten a rope by manipulating it so that both ends are placed into distinct target areas.

\textbf{Success Metric:} The task is considered successfully completed when both ends of the rope are positioned within their respective target areas.

\textbf{Objects:} The task involves a single object: a rope.

\textbf{Coordination Challenges:} The main challenge involves handling a deformable object, which requires the robot to grasp and manipulate the rope simultaneously at different points to achieve the desired straightening.

\textbf{Language Instructions:} \textit{Straighten the rope.}

\subsection*{(j) sweep dust pan}

\begin{wrapfigure}{l}[0cm]{0.4\textwidth}
    \centering
    \vspace{-\normalbaselineskip}
    \includegraphics[width=.38\textwidth,keepaspectratio=true]{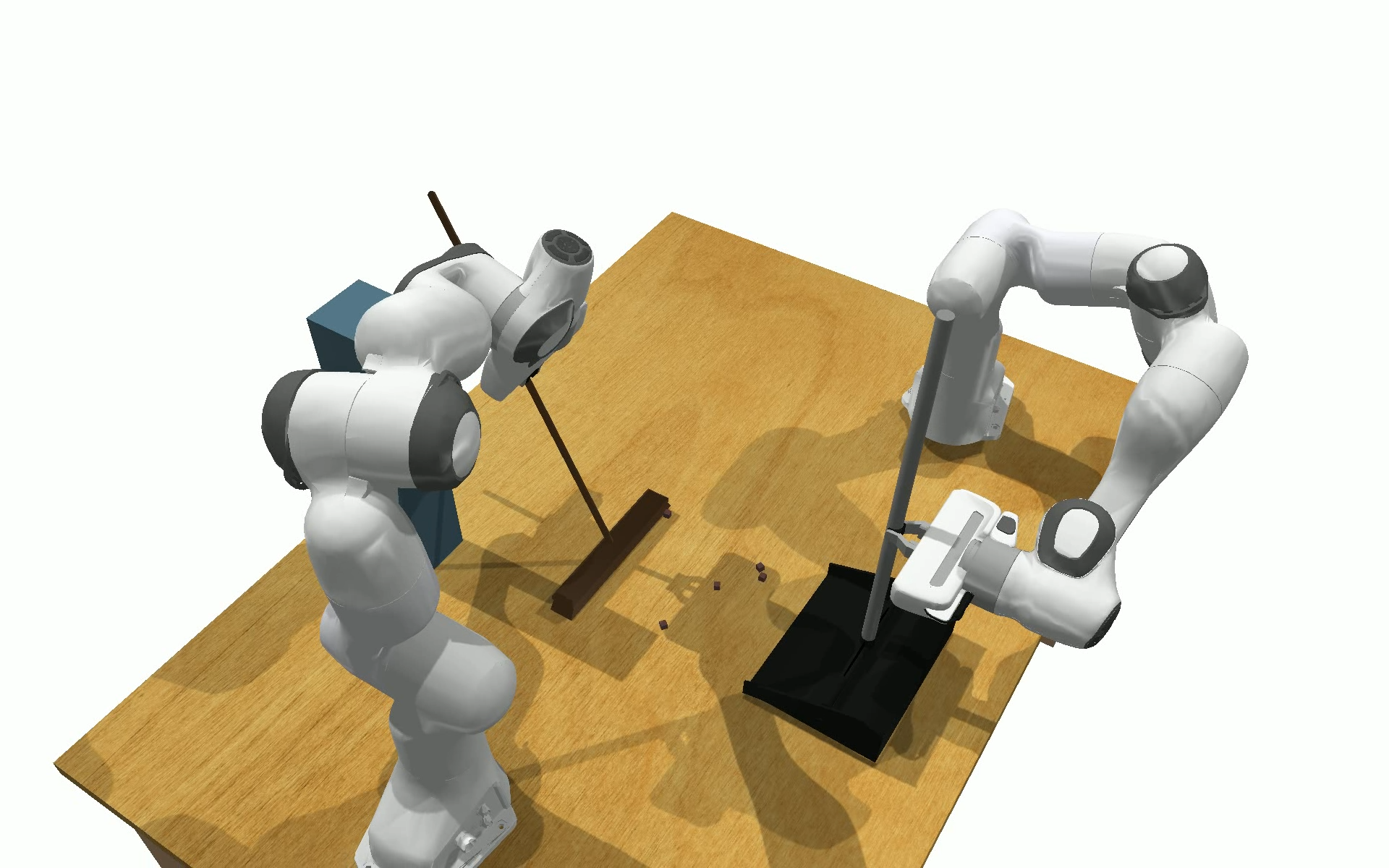}
    \vspace{-\normalbaselineskip}
\end{wrapfigure}

\textbf{Task Description:} The robot's task is to sweep the dust into the dust pan using a broom. This involves coordinating the sweeping motion to ensure the dust is effectively collected.

\textbf{Success Metric:} The task is considered successfully completed when all the dust is inside the dust pan.

\textbf{Objects:} The task involves several objects: a broom, a dust pan, supporting objects, and dust.

\textbf{Coordination Challenges:} The main challenge lies in executing the sweeping motion accurately to ensure that the dust is directed into the dust pan.

\textbf{Language Instructions:} \textit{Sweep the dust to the pan.}

\subsection*{(k) lift tray}

\begin{wrapfigure}{l}[0cm]{0.4\textwidth}
    \centering
    \vspace{-\normalbaselineskip}
    \includegraphics[width=.38\textwidth,keepaspectratio=true]{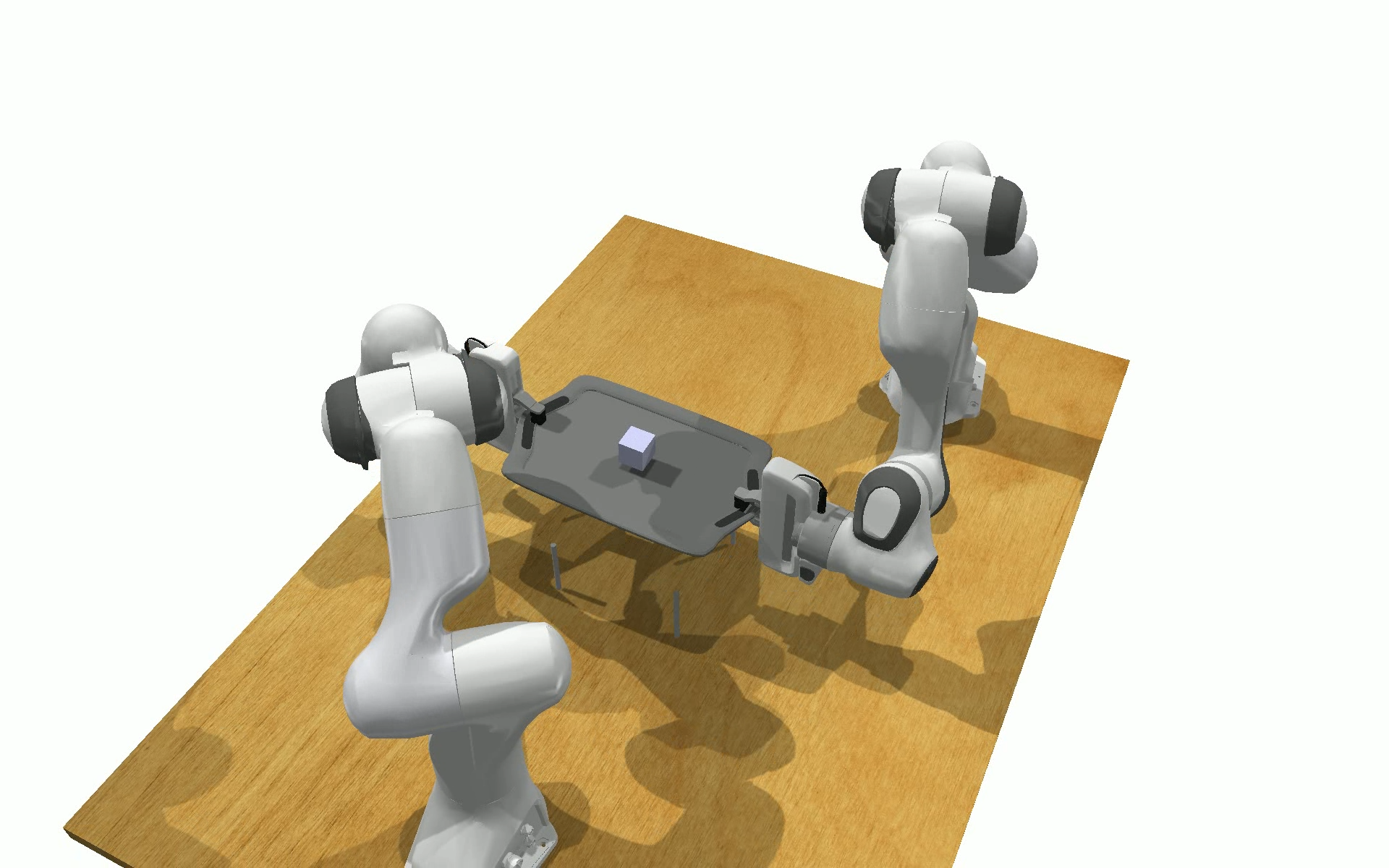}
    \vspace{-\normalbaselineskip}
\end{wrapfigure}

\textbf{Task Description:} The robot's task is to lift a tray that is placed on a holder. An item is on top of the tray and must be balanced while both arms lift the tray.

\textbf{Success Metric:} The task is considered successfully completed when both the tray and the item on top reach a height above \SI{1.2}{\meter}.

\textbf{Objects:} The task involves a tray, a holder, and an item.

\textbf{Coordination Challenges:} The primary challenge lies in coordinating the lifting motion with both arms to maintain the balance of the item on the tray. This task cannot be accomplished with only one arm.

\textbf{Language Instructions:} \textit{Lift the tray}

\subsection*{(l) handover item (easy)}

\begin{wrapfigure}{l}[0cm]{0.4\textwidth}
    \centering
    \vspace{-\normalbaselineskip}
    \includegraphics[width=.38\textwidth]{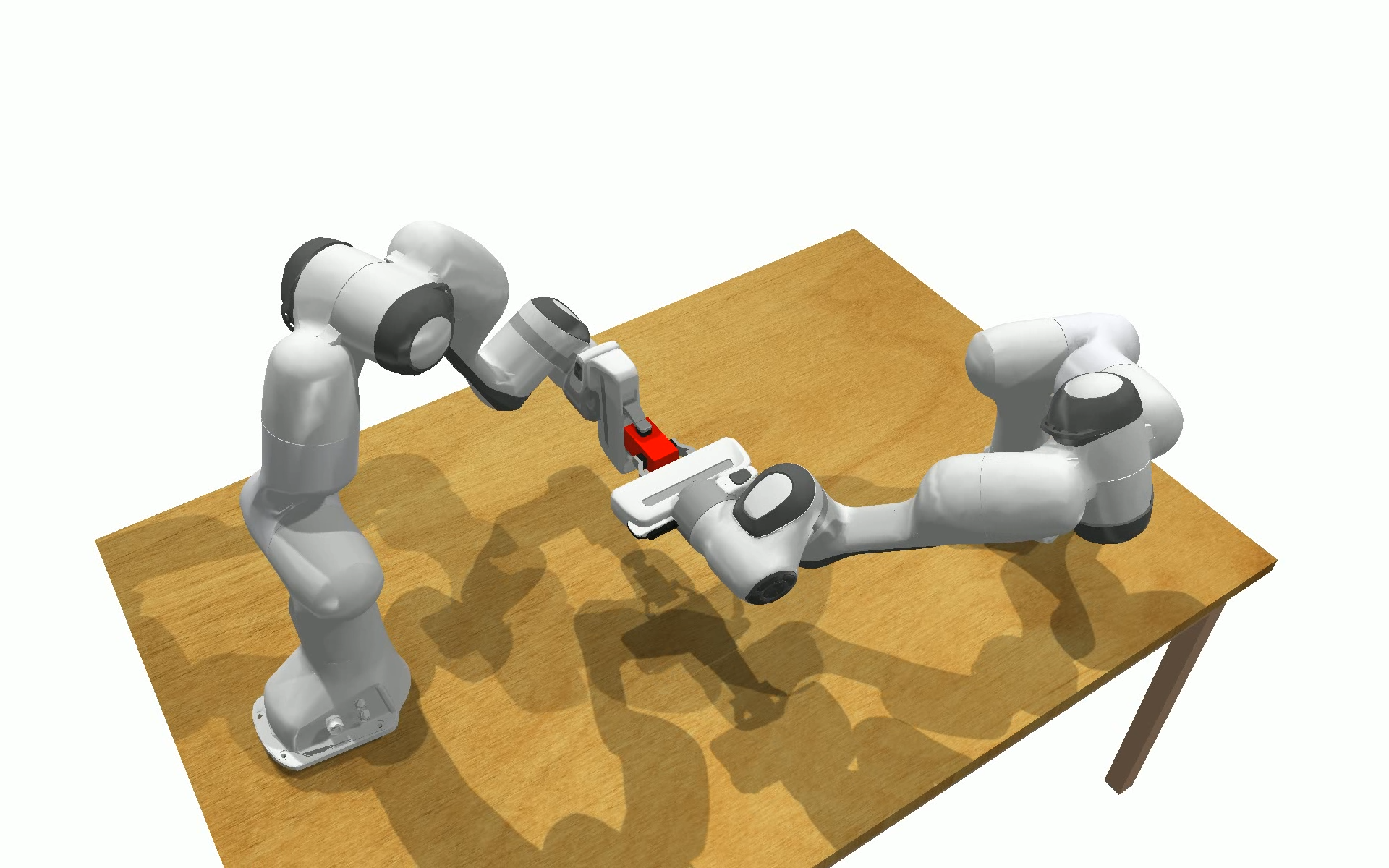}
    \vspace{-\normalbaselineskip}
\end{wrapfigure}

\textbf{Task Description:} The robot's task is to hand over a red item. One robotic arm must grasp the red item while the other arm remains free and wait for the handover

\textbf{Success Metric:} The task is considered successfully completed when the robot has securely grasped the correct item and lifted it to a height of \SI{80}{\centi\meter}, while the other arm remains idle and has not grasped anything.

\textbf{Objects:} The task involves a red block.

\textbf{Coordination Challenges:} The primary challenge lies in coordinating the handover process.

\textbf{NB:} There is also a more complex variant of this task that involves handling multiple objects of different shapes and sizes. Refer to the "handover item" task for details.

\textbf{Language Instructions:} \textit{Handover the item.}

\subsection*{(m) take tray out of oven}

\begin{wrapfigure}{l}[0cm]{0.4\textwidth}
    \centering
    \vspace{-\normalbaselineskip}
    \includegraphics[width=.38\textwidth,keepaspectratio=true]{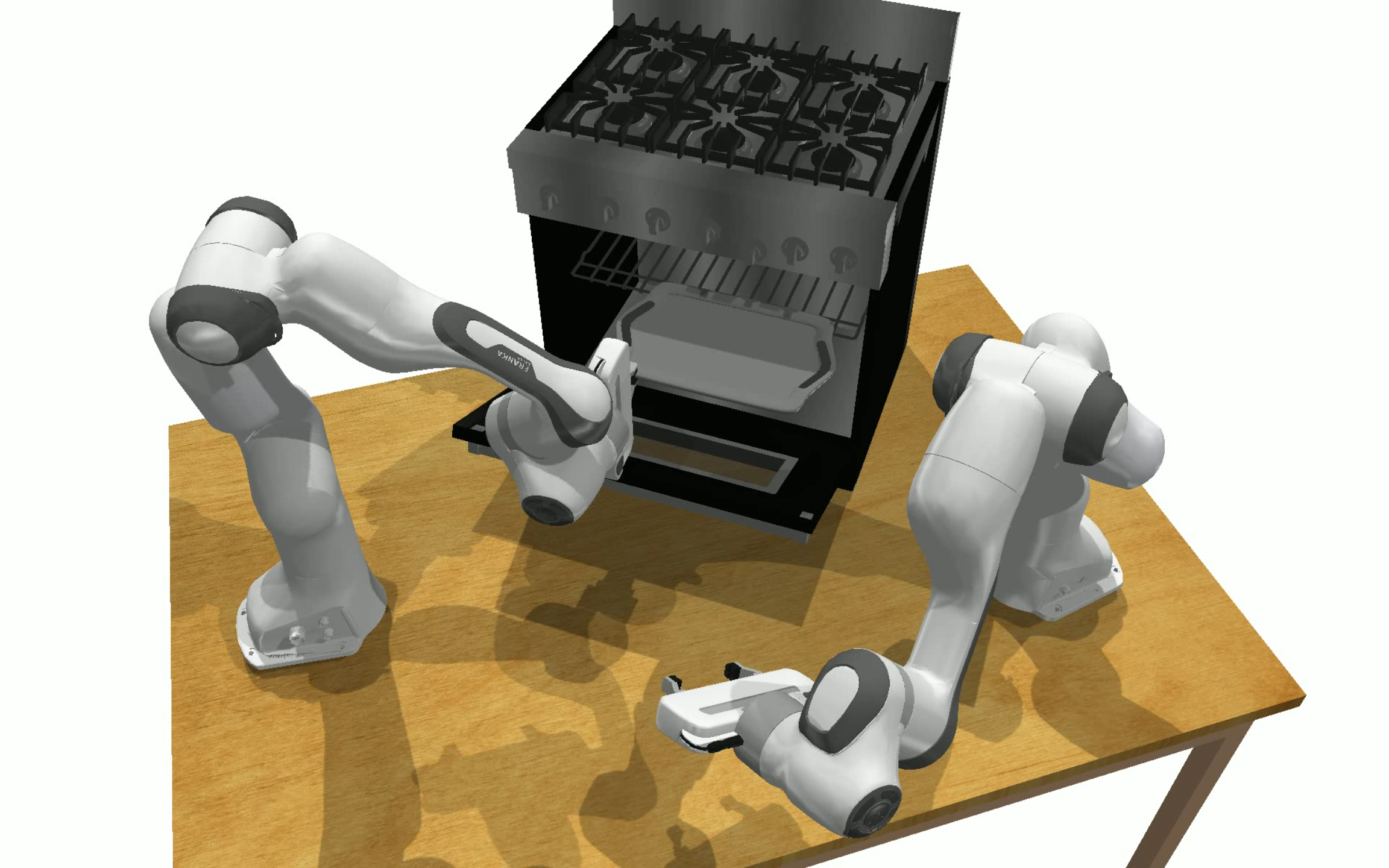}
    \vspace{-\normalbaselineskip}
\end{wrapfigure}

\textbf{Task Description:} The robot's task is to remove a tray that is located inside an oven. This involves opening the oven door and then grasping the tray.

\textbf{Success Metric:} The task is considered successfully completed when the tray is lifted above the oven.

\textbf{Objects:} The task involves a tray inside an oven.

\textbf{Coordination Challenges:} The primary challenge lies in opening the oven door to make the tray graspable.

\textbf{NB:} This task can be solved with only one arm.

\textbf{Language Instructions:} \textit{Take tray out of oven.}

\end{document}